\documentclass{article}
\usepackage{arxiv}
\usepackage[noadjust]{cite}
\usepackage[utf8]{inputenc} 
\usepackage[T1]{fontenc}    
\usepackage[colorlinks,urlcolor=blue,linkcolor=blue,citecolor=blue]{hyperref}      
\usepackage{url}            
\usepackage{booktabs}       
\usepackage{amsfonts}       
\usepackage{nicefrac}       
\usepackage{microtype}      
\usepackage{xcolor}  
\usepackage{subfigure}
\usepackage{multirow}
\usepackage{graphicx}
\usepackage{subfigure}
\usepackage{mathtools}
\usepackage{bm}
\usepackage{algpseudocode,algorithm}
\usepackage{cleveref}

\newtheorem{prop}{Proposition}

\newcommand{\bu}{\bm{u}}
\newcommand{\btheta}{\bm{\theta}}

\newcommand{\blambda}{\bm{\lambda}}
\newcommand{\bmu}{\bm{\mu}}
\newcommand{\N}{\bm{\mathcal{N}}}
\newcommand{\M}{\bm{\mathcal{M}}}
\newcommand{\bH}{\bm{\mathcal{H}}}
\newcommand{\f}{\bm{f}}

\renewcommand{\shorttitle}{}

\title{A memory-efficient neural ordinary differential equation framework based on high-level adjoint differentiation}
\author{Hong Zhang \thanks{This material is based upon work supported by the U.S. Department of
Energy, Office of Science, Office of Advanced Scientific Computing Research (ASCR),
Scientific Discovery through Advanced Computing (SciDAC) program through the
FASTMath Institute under contract DE-AC02-06CH11357 at Argonne National
Laboratory and through the ASCR DOE-FOA-2493 “Data-intensive scientific machine learning”. (\textit{Corresponding author: Hong Zhang.})}
\\
Division of Mathematics and Computer Science\\
Argonne National Laboratory \\
\texttt{hongzhang@anl.gov}
\And
Wenjun Zhao\\
Department of Mathematics\\
Brown University \\
\texttt{wenjun\_zhao@brown.edu}
}

\begin{document}

\maketitle

\begin{abstract}
Neural ordinary differential equations (neural ODEs) have emerged as a novel
network architecture that bridges dynamical systems and deep learning. However,
the gradient obtained with the continuous adjoint method in the vanilla neural
ODE is not reverse-accurate. Other approaches suffer either from an excessive
memory requirement due to deep computational graphs or from limited choices for
the time integration scheme, hampering their application to large-scale complex
dynamical systems. To achieve accurate gradients without compromising memory
efficiency and flexibility, we present a new neural ODE framework, PNODE, based
on high-level discrete adjoint algorithmic differentiation. By leveraging
discrete adjoint time integrators and advanced checkpointing strategies tailored
for these integrators, PNODE can provide a balance between memory and
computational costs, while computing the gradients consistently and accurately.
We provide an open-source implementation based on PyTorch and PETSc, one of the
most commonly used portable, scalable scientific computing libraries. We
demonstrate the performance through extensive numerical experiments on image
classification and continuous normalizing flow problems. We show that PNODE
achieves the highest memory efficiency when compared with other reverse-accurate
methods. On the image classification problems, PNODE is up to two times faster
than the vanilla neural ODE and up to $2.3$ times faster than the best existing
reverse-accurate method. We also show that PNODE enables the use of the implicit
time integration methods that are needed for stiff dynamical systems.
\end{abstract}

\keywords{
  Neural ODEs, deep learning, adjoint differentiation, checkpointing, implicit methods}

\section{Introduction}

A residual network can be seen as a forward Euler discretization
of a continuous time-dependent ordinary differential equation (ODE), as
discussed in \cite{Lu2018,Haber_2017,Ruthotto2018}. In the limit of smaller time
steps, a new family of deep neural network models called neural ODEs was
introduced in \cite{NODE}. Compared with traditional discrete-time models, this
family is advantageous in a wide range of applications, such as modeling of
invertible normalizing flows \cite{grathwohl2018scalable} and continuous time
series with irregular observational data \cite{Rubanova2019}. This continuous
model also makes neural ODEs particularly attractive for learning and modeling
the nonlinear dynamics of complex systems. Neural ODEs have been successfully
incorporated into many data-driven models
\cite{Rubanova2019,GreydanusGoogleBrain2019,ayed2019learning} and adapted to a
variety of differential equations including hybrid systems \cite{Dupont2019} and
stochastic differential equations \cite{Jia2019}.

The success of neural ODEs consolidates the connection between deep learning and
dynamical systems, allowing well-established dynamical system theory to be
applied to deep learning. However, training neural ODEs efficiently is still
challenging. A major part of the challenge is computing the gradient for the ODE
layer, while achieving a balance between \textit{stability}, \textit{accuracy},
and \textit{memory efficiency}. A naive approach for the gradient calculation is
to \textit{backpropagate} the ODE layer with low-level automatic differentiation
(AD) directly, resulting in a large, redundant computational graph. To overcome
this memory limitation, Chen et al.~\cite{NODE} proposed using a continuous
adjoint method in place of backpropagation and reverse the forward trajectory,
which requires no storage of previous states and allows for training with
constant memory. As pointed out in
\cite{ANODE,onken2020discretizeoptimize,zhuang2020}, however, the continuous
adjoint method may lead to inaccurate gradients and instability during training.
Therefore, reverse-accurate methods based on discrete adjoint methods and
backpropagation with low-level AD have been developed in \cite{ANODE, ANODEV2,
zhuang2020, zhuang2021mali}, and certain simple checkpointing strategies are
used to reduce the memory overhead. The MALI method in \cite{zhuang2021mali} can
avoid the use of checkpointing, but it is restricted to a symplectic time
integrator that is typically designed for Hamiltonian systems.

Another limitation of existing frameworks is that they support only explicit
time integration methods such as Runge--Kutta methods. This leads to tremendous
difficulties in dealing with stiff dynamical systems, partly because of the
stability constraints of explicit schemes. Implicit schemes are unconditionally
stable; however, they require the solution of nonlinear/linear systems. Directly
backpropagating through implicit schemes with low-level AD is not
computationally efficient and can be infeasible because of the complexity in the
procedure, which is usually iterative \cite{Beck1994}.

Therefore, we propose PNODE to overcome the limitations of existing neural ODEs.
By utilizing a high-level discrete adjoint method together with checkpointing,
PNODE achieves reverse accuracy and memory efficiency in the gradient
calculations and, at the same time, allows for flexibility in training
strategies. The main contributions of this work are as follows:
\begin{enumerate}
\item A framework that minimizes the depth of the computational graph for neural
network backpropagation, leading to significant savings in memory for neural
ODEs. Our code is available online.\footnote{\href{https://github.com/caidao22/pnode}{https://github.com/caidao22/pnode}}
\item High-level discrete adjoint calculation combined with neural network
backpropagation that allows more flexibility in the design of neural ODEs. We
show that our framework enables the use of implicit time integration schemes and
opens up the possibility of incorporating other integration methods.
\item Demonstration that PNODE outperforms existing methods in memory efficiency
on diverse tasks including image classification and continuous normalizing flow
problems.
\item Successful application of PNODE to learning stiff dynamical systems with
implicit methods, which has been recognized to be very challenging.
\end{enumerate}

The rest of this article is organized as follows. In Section \ref{sec:prelim} we
introduce the basics of neural ODEs and discuss the two adjoint methods that are
commonly used techniques for computing gradients when solving optimal control
problems. In Section \ref{sec:method} we present the key techniques in our
discrete adjoint-based framework, explain their advantages over existing
frameworks, and describe the implementation. In Section \ref{sec:exp} we
demonstrate the performance of our approach and showcase its success in learning
stiff dynamics from data. In Section \ref{sec:conclusion} we summarize our
conclusions.

\section{Preliminaries}\label{sec:prelim}

\subsection{Neural ODEs as an optimal control problem}
Neural ODEs are a class of models that consider the continuum limit of neural
networks where the input-output mapping is realized by solving a system of
parameterized ODEs:
\begin{equation}
    \frac{d \bu}{d t} = \f(\bu, \btheta, t) \quad \bu(t_0) = \bu_0, \quad t \in [t_0,t_F],
    \label{eq:ode}
\end{equation}
where $\bu \in \mathbb{R}^N$ is the state, $\btheta \in \mathbb{R}^{N_p}$ are
the weights, and $f: \mathbb{R}^N \times \mathbb{R}^{N_p} \times \mathbb{R}
\rightarrow \mathbb{R}^N$ is the vector field approximated by a neural network
(NN). The input-output mapping is learned through a data-driven approach. During
training, $\bu^{(1)}, \bu^{(2)}, \dots, \bu^{(S)}$ are input as initial states,
and the output should match the solutions of \eqref{eq:ode} at $t_F$.

Training neural ODEs can be viewed as solving an optimal control problem,
\begin{equation}
\begin{aligned}
    & \min_{\theta,\bu} \left\{ \mathcal{L} \coloneqq \phi(\bu(t_F)) + \int_{t_0}^{t_F} q(\bu(t),t) dt \right\},\\
    & \text{subject to the dynamical constraint \eqref{eq:ode}.}
\end{aligned}
\label{eq:optimal_control}
\end{equation}
This formulation generalizes the loss functional that depends on the final
solution in the vanilla neural ODEs to additional functionals that depend on the
entire trajectory in the time (or depth) domain. The integral term may come as
part of the loss function or as a regularization term, such as Tikhonov
regularization, which is typically used for solving ill-posed problems. Finlay
et al.~\cite{Finlay2020} showed that regularization techniques can be used to
significantly accelerate the training of neural ODEs. Their approach can also be
captured by the formula \eqref{eq:optimal_control}.

\textbf{Notation} For convenience, we summarize the notation used in the
article:
\begin{itemize}
\setlength\itemsep{0em}
    \item $N_t$: number of time steps in time integration
    \item $N_t^B$: number of time steps in a backward pass
    \item $N_s$: number of stages in the time integration method
    \item $N_l$: number of layers in the NN that approximates the vector field $\f$ in \eqref{eq:ode}.
    \item $N_b$: number of ODE blocks (one instance of \eqref{eq:ode} corresponds to one block).
    \item $N_c$: number of maximum allowed checkpoints
\end{itemize}

\subsection{Discrete adjoint vs continuous adjoint} \label{sec:disadjoint}

In order to evaluate the objective $\mathcal{L}$ in \eqref{eq:optimal_control},
both the ODEs \eqref{eq:ode} and the integral need to be discretized in time. In
order to calculate the gradient of $\mathcal{L}$ for training, adjoint methods
(i.e., backpropagation in machine learning) can be used. Two distinct ways of
deriving the adjoints exist:  continuous adjoint and discrete adjoint.

\textbf{Continuous adjoint}
The continuous adjoint sensitivity equation is
\begin{eqnarray}
\frac{d \widetilde{\blambda}}{d t} & = &- \left(\frac{\partial \f}{\partial \bu}\right)^T \widetilde{\blambda} - \frac{\partial q}{\partial \bu}, \\
\widetilde{\blambda}(t_F) & = & \frac{\partial \mathcal{L} }{\partial \bu(t_F)},
\end{eqnarray}
where $\widetilde{\blambda} \in \mathbb{R}^N$ is an adjoint variable. Its
solution gives the gradient of $\mathcal{L}$ with respect to the initial state
$\bu_0$. That is, $\widetilde{\blambda}(t_0) = \frac{d \mathcal{L}}{d \bu_0}$.
The gradient of $\mathcal{L}$ with respect to the weights $\btheta$ is
\begin{equation}
    \frac{d \mathcal{L} }{d \btheta} = \frac{\partial \mathcal{L} }{\partial \btheta} + \int_{t_0}^{t_F} \left( \widetilde{\blambda}^T \frac{\partial \f}{\partial \btheta} + \frac{\partial q}{\partial \btheta} \right) dt.
\end{equation}
The infinite-dimensional sensitivity equations and the integrals also need to be
discretized in time in order to obtain a finite-dimensional approximation to the
derivatives. In a practical implementation, one must integrate \eqref{eq:ode}
forward in time and save the intermediate states. Then the sensitivity equations
are solved backward in time, during which the saved states are restored to
evaluate the Jacobians. These two steps are analogous to forward pass and
backpropagation in the machine learning nomenclature. Theoretically, different
time integration methods and different time step sizes can be chosen for solving
the forward equations and the sensitivity equations.

\textbf{Discrete adjoint} An alternative approach is to derive the adjoint for
the discretized version of the infinite-dimensional equations \eqref{eq:ode}
directly. Assume the discretization is formally represented by a time-stepping
operator $\N$ that propagates the ODE solution from one time step to another:
\begin{equation}
    \bu_{n+1} = \N (\bu_n, \btheta).
    \label{eq:ts_alg}
\end{equation}
The adjoint sensitivities are propagated by
\begin{equation}
  \begin{aligned}
  \blambda_{n} &=  \left(\frac{\partial \N}{\partial \bu}(\bu_n)\right)^T \blambda_{n+1},  \\
  \bmu_{n} &=  \left(\frac{\partial \N}{\partial \theta}(\bu_n)\right)^T \blambda_{n} + \bmu_{n+1}, \quad n= N-1, \dots, 0, \\
  \end{aligned}
\label{eqn:disadjoint}
\end{equation}
with the terminal condition
\begin{equation}
\blambda_{N} = \left(\frac{\partial \mathcal{L}}{\partial \bu_N}\right)^T, \quad \bmu_{N} = \left(\frac{\partial \mathcal{L}}{\partial \btheta}\right)^T.
\label{eqn:disadjoint_tc}
\end{equation}
The solutions $\blambda_{0}$ and $\bmu_{0}$ represent the sensitivity (also
called gradient) of $\mathcal{L}$ to the initial state $\bu_0$ and to the
parameters $\btheta$, respectively. Given a time integration scheme, one can
derive a specific discrete adjoint formula in the form \eqref{eqn:disadjoint}.
For example, the discrete adjoint for the forward Euler method is listed in
Table \ref{tab:adjoint_feuler}. We refer readers to
\cite{Zhang2022tsadjoint,Zhang2014} for more examples and the details on their
derivation.

\subsection{Choice of adjoint method for neural ODEs}

While both approaches can generate the gradients for the loss function, these
gradients are typically not the same even if the same time integration method
and step sizes are used. They are asymptotically equivalent as the time step
size approaches zero. As a concrete example, we compare the continuous adjoint
equation after time discretization using the forward Euler method with the
discrete adjoint for the same method. As shown in Table
\ref{tab:adjoint_feuler}, the Jacobian is evaluated at $\bu_{n+1}$ and $\bu_n$,
respectively, causing a discrepancy that can be bounded by Proposition
\ref{prop:1}.
\begin{table}[ht]
\centering
\caption{Comparison of different adjoint approaches for forward Euler.}
\label{tab:adjoint_feuler}
\tabcolsep=0.08cm
\begin{tabular}{l c}
\toprule
forward propagation & $\displaystyle \bu_{n+1} = \bu_{n} + h \f(\bu_{n},\btheta,t_n)$\\
continuous adjoint & $\displaystyle \widetilde{\blambda}_{n} = \widetilde{\blambda}_{n+1} + h \left( \frac{\partial \f(\bu_{n+1},\btheta,t_{n+1})}{\partial \bu} \right)^T \widetilde{\blambda}_{n+1} $  \\
\rule{0pt}{4ex}discrete adjoint &  $\displaystyle \blambda_{n} = \blambda_{n+1} + h \left( \frac{\partial \f(\bu_{n},\btheta,t_n)}{\partial \bu} \right)^T \blambda_{n+1} $ \\
\bottomrule
\end{tabular}
\end{table}

\begin{prop}
Assuming $\widetilde{\blambda}_{n+1} = \blambda_{n+1}$, the local discrepancy
between the continuous adjoint sensitivity and the discrete adjoint sensitivity
is
\begin{equation}
    \| \widetilde{\blambda}_{n}  - \blambda_{n} \| \leq   h^2  \| \bH\left(\bu_n + \epsilon(\bu_{n+1} - \bu_n),\btheta,t_n\right) \f(\bu_n,\btheta,t_n) \| \| \blambda_{n+1} \|,
\end{equation}
where $\bH$ is the Hessian of $\f$ and $\epsilon$ is a constant in $(0,1)$.
\label{prop:1}
\end{prop}

This proposition can be easily proved by using Taylor's expansion and the
triangle inequality for the norm. One can see that $\widetilde{\blambda}_{n} =
\blambda_n$ for linear functions since the Hessian is zero and for nonlinear
functions, such as deep neural networks, $\widetilde{\blambda}_{n}$ approaches
$\blambda_n$ quadratically as $h \rightarrow 0$. Nevertheless, the accumulated
discrepancy may clearly lead to inconsistency in the gradient calculated with
the continuous adjoint approach. Similar results also hold for other multistage
time integration methods such as Runge--Kutta methods. Many studies
\cite{Nielsen2006,Nielsen2013,Zhang2017} have suggested that the discrete
adjoint approach produces accurate gradients that match the machine precision,
making it more favorable for gradient-based optimization algorithms.

\subsection{High-level abstraction of automatic differentiation} \label{sec:ad}

Despite the nuances in the different approaches for the gradient computations in
neural ODEs, they can be interpreted fundamentally as a unified method that
utilizes automatic differentiation \cite{NoceWrig06} at different abstraction
levels. From the theoretical perspective, they all can be viewed as the reverse
(adjoint) mode of AD, which can compute the gradient of a scalar differentiable
function with respect to many parameters efficiently because its computational
cost is independent of the number of parameters. The reverse mode of AD
comprises two steps: a forward pass to generate output from the input and a
backward pass to compute the derivatives. Many AD tools build
\textit{computational graphs} for convenience to express complex models in the
forward pass. Once the graph is defined using \textit{primitive} operations,
derivatives can be calculated automatically through the chain rule of
differentiation based on ``local'' derivatives of these operations.

The \textit{primitive} operation can be defined at different abstraction levels
in different approaches. The backpropagation algorithm used in popular machine
learning (ML) platforms such as \texttt{TensorFlow} and \texttt{PyTorch} take
the functions on tensors as the primitive operations. In the continuous adjoint
approach adopted in vanilla neural ODEs, the primitive operation is the ODE
itself; in the discrete adjoint approach in Section~\ref{sec:disadjoint}, the
primitive operations are the functions on the states $\bu$ (including
intermediate stage values) in \eqref{eq:ode}, while the propagation equation
\eqref{eq:ts_alg} for a particular time-stepping algorithm consists of a
sequence of such primitive operations, thus yielding a high-level representation
of the computational procedure. Computational graphs may also be used in these
high-level adjoint approaches, but it is often possible to derive and implement
the adjoint method manually without using
graphs~\cite{Zhang2022tsadjoint,Zhang2014}.

\section{Method}\label{sec:method}

In this section we first explain the advantage of high-level AD approaches over
a naive approach for the derivative calculation of neural ODEs. Next, based on
the high-level AD approach, we propose a new neural ODE framework, PNODE. We
show how PNODE enables implicit time integration, which has not been achieved
with other neural ODE approaches. We then describe the implementation of PNODE
based on \texttt{PETSc}.

\subsection{Memory consumption for AD}\label{sec:naive}

In a naive neural ODE approach, one can use an AD tool such as AutoGrad to
backpropagate through the entire forward ODE solve. This approach requires
recording a deep computational graph that consists of all the primitive
operations (for example, tensor operations in \texttt{PyTorch}). The size of the
graph is proportional to the size of the NN and the number of time steps; thus,
the memory cost for backpropagation is $\mathcal{O}(N_t N_s N_l)$. This makes
the approach infeasible for large problems and long-time time integration.

In contrast, high-level AD methods, such as the discrete adjoint method we
discussed in Section \ref{sec:disadjoint}, compute the gradient by composing the
derivatives for high-level primitive operations. Therefore, one can
differentiate through only the primitive operation instead of through the entire
ODE solver. In the discrete adjoint method, the derivatives for most primitive
operations are straightforward to obtain. The only difficulty is to compute the
derivative (also called Jacobian) for the function $f$. However, one can use the
backpropagation algorithm since $f$ is approximated by an NN. Note that the
computational graph for backpropagation will be much shallower compared with the
naive approach, and thus the memory cost of backpropagation is independent of
the number of time steps. In addition, adjoint methods require accessing the
intermediate states to solve the adjoint equations, as we mentioned in Section
\ref{sec:disadjoint}. These states can be obtained with a \textit{checkpointing}
technique. The memory cost for checkpointing is dependent of the number of time
steps but is independent of the size of the NN.
\begin{figure}
  \centering
  \includegraphics[width=0.6\linewidth]{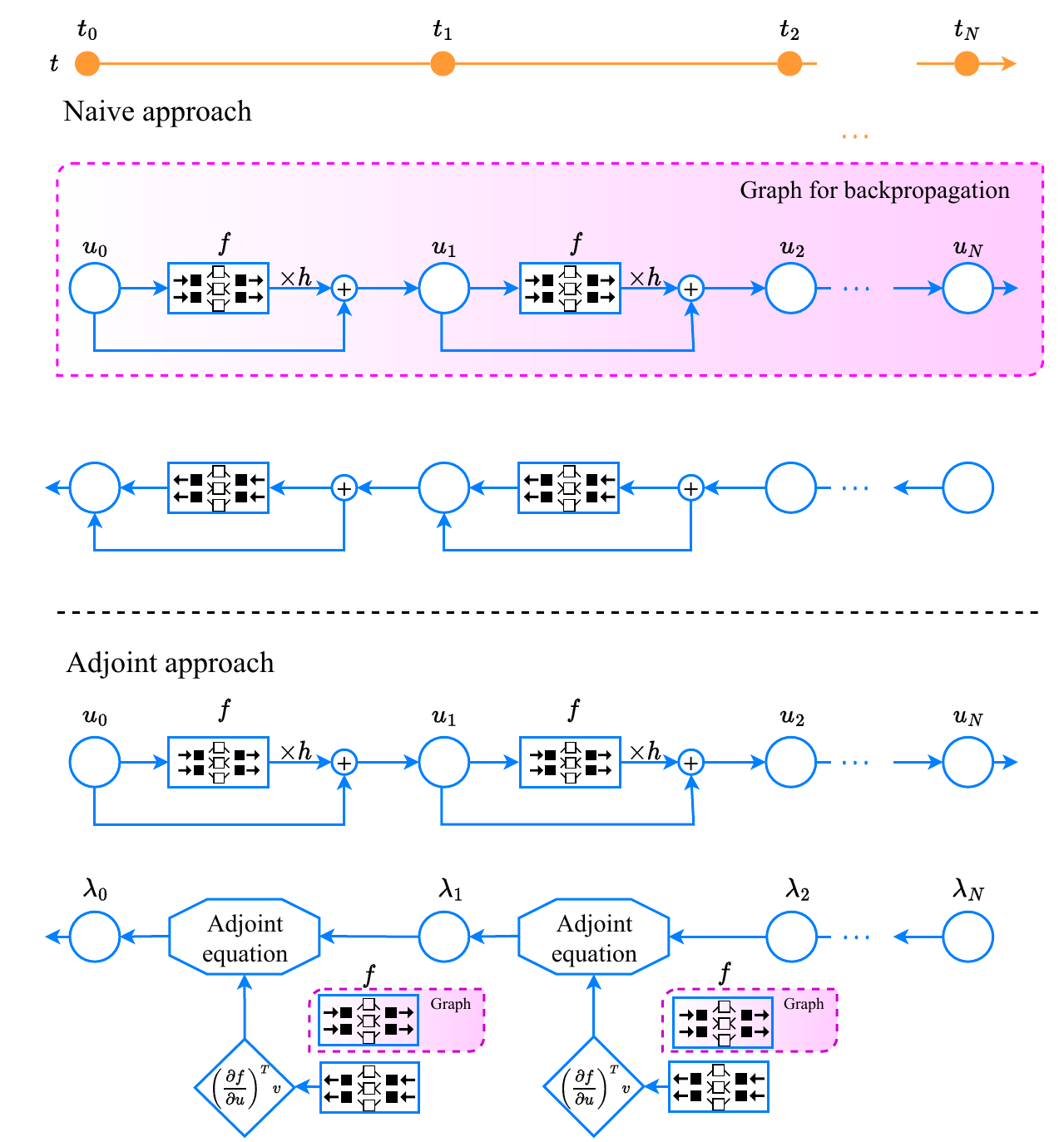}
  \caption{Schematic illustrating how the adjoint approach leads to shallower graph for backpropagation compared with the naive approach. The forward model corresponds to the Euler method.}
  \label{fig:graph}
\end{figure}

\subsection{Proposed PNODE framework}

Based on the discrete adjoint method, we propose a new neural ODE framework,
PNODE, that achieves reverse accuracy, memory efficiency, and flexibility at the
same time. Unlike the naive approach, PNODE does not record any computational
graph in the forward pass. Instead, we checkpoint the solutions at select time
steps, with the stage values if multistage time integration methods such as
Runge--Kutta methods are used. In the reverse pass, the checkpoints are restored
and used to compute $\left(\frac{\partial \f}{\partial \bu} \right)^T $ or the
transposed Jacobian-vector product $\left(\frac{\partial \f}{\partial \bu}
\right)^T v $, where the vector $v$ is determined by the time integration
algorithm. The transposed Jacobian-vector product is a required ingredient in
the discrete adjoint method; it is obtained by backpropagating $\f$ with a
constant memory cost $\mathcal{O}(N_l)$, as illustrated in Figure
\ref{fig:graph}. Note that the high-level discrete adjoint formula
\eqref{eqn:disadjoint} can be derived for any time integration method and
implemented manually, and it can be reused in different applications. Automatic
derivation and implementation of the adjoint model might be possible with
similar techniques used by the dolfin-adjoint project \cite{Farrell2013}. The
algorithm for training PNODE is summarized in Algorithm \ref{eq:ts_alg}.

\begin{algorithm}
  \caption{PNODE to achieve accuracy and memory efficiency with high-level adjoint differentiation.}
  \textbf{Forward}
  \begin{algorithmic}
  \State Solve \eqref{eq:ode} with a time-stepping algorithm \eqref{eq:ts_alg}
  \State Checkpoint the states and stage values for all or selective time steps
  \State Compute the loss $\mathcal{L}$
  \end{algorithmic}
  \textbf{Backward}
  \begin{algorithmic}
  \State Initialize the adjoint variables $\blambda$ and $\bmu$ with \eqref{eqn:disadjoint_tc}
  \For{$n \coloneqq N-1$ to $0$}
      \State Restore from the closest checkpoint, and recompute the forward pass to time $t_{n+1}$
      \State Compute $\blambda_n$ and $\bmu_n$ with \eqref{eqn:disadjoint}
  \EndFor
  \State Output the gradient $\frac{d \mathcal{L} }{d \btheta} = \mu_0$
  \end{algorithmic}
\end{algorithm}

The checkpointing process in PNODE provides a trade-off between storage and
computational overhead. The maximum space needed for checkpointing is
$(N_t\text{-}1) \times size\_of\_a\_checkpoint$, where a checkpoint consists of
the state vector and $N_s$ stage vectors for one time step. In the ideal case
where  memory is sufficient for saving all stages, no time steps are recomputed
in the backward pass. With a limited memory budget ($N_c < N_t{-}1$), the states
that are not checkpointed in the forward pass can be recomputed from a nearby
checkpoint. To minimize the number of recomputations, we use the binomial
strategy presented in \cite{Zhang2021iccs,Zhang2022TOMS}. The algorithm is an
extension of the classic \texttt{Revolve} algorithm \cite{checkpointing2}. Its
optimality for multistage time integration methods is proved in
\cite{Zhang2021iccs} and summarized below. We refer readers to
\cite{Zhang2021iccs,Zhang2022TOMS} for illustrations of the checkpointing
procedure.

\begin{prop} \cite{Zhang2021iccs,Zhang2022TOMS} Given $N_c$ allowed checkpoints
in memory, the minimal number of extra forward steps (recomputations) needed for
the adjoint computation of $N_t$ time steps is
\begin{equation}
\tilde{p}(N_t,N_c)= (t-1)\, N_t-\binom{N_c+t}{t-1} + 1,
\label{eqn:ptilde}
\end{equation}
where $t$ is the unique integer satisfying $\binom{N_c+t-1}{t-1} < N_t \leq
\binom{N_c+t}{t}$.
\label{prop:recomp}
\end{prop}

The total memory cost of PNODE thus consists of two parts: backpropogation for
NN (the function $\f$) and checkpointing for the adjoint calculation. NN
backpropagation requires $\mathcal{O}(N_l)$ memory, while the memory cost of
checkpointing is at most $\mathcal{O}((N_t\text{-}1) (N_s\text{+}1))$.

The computation cost of PNODE can be split into three parts:  the forward
computation, the reverse computation, and the recomputation overhead. The
forward or reverse computation cost is $\mathcal{O}(N_t N_l)$, which is common
for any neural ODE approach. The recomputation cost depends on the memory budget
and becomes zero when $(N_t\text{-}1)$ checkpoints are allowed.

\subsection{Enabling implicit time integration}

The application of implicit time integration is highly desired for stiff
dynamical systems where explicit methods may fail because of stability
constraints. Backpropagating through the implicit solver with a low-level AD
tool is difficult, however, because of the complexity of the nonlinear/linear
solve required at each time step and the large amount of memory needed for the
resulting computational graph. By taking the function ($\f$) evaluation as a
\textit{primitive} operation in high-level AD, PNODE excludes the
nonlinear/linear solvers from the computational graph for backpropagation.
Instead, it solves an adjoint equation (a transposed linear system) to propagate
the gradients. Take the backward Euler method for example. Consider a more
general parameterized dynamical system in the form (common for mechanical
systems)
\begin{equation}
  \M \frac{d \bu}{d t} = \f(\bu, \btheta, t) \quad \bu(t_0) = \bu_0, \quad t \in [t_0,t_F],
  \label{eq:ode_mass}
\end{equation}
where $\M$ is the mass matrix (when $\M$ is an identity matrix, it falls back to
\eqref{eq:ode}). The backward Euler formula for solving this system is
\begin{equation}
  \M \bu_{n+1} = \M \bu_n + h_n \f(\bu_{n+1},\btheta,t_{n+1}).
\label{eqn:beuler}
\end{equation}
Its discrete adjoint in a simplified form (with no integral term in loss
function \eqref{eq:optimal_control}) can be written as
\begin{equation}
  \label{eqn:disadj_beuler_full}
\begin{aligned}
  \M^T\blambda_s &= \blambda_{n+1} + h\,\left(\frac{\partial \f(\bu_{n+1},\btheta,t_{n+1}}{\partial \bu}\right)^T\,\blambda_s, \\
  \blambda_n &= \blambda_{n+1} + h\,\left(\frac{\partial \f(\bu_{n+1},\btheta,t_{n+1}}{\partial \bu}\right)^T\,\blambda_s ,\\
  \bmu_n &= \bmu_{n+1} + h\,\left(\frac{\partial \f(\bu_{n+1},\btheta,t_{n+1}}{\partial \btheta}\right)^T\,\blambda_s .
\end{aligned}
\end{equation}
At each reverse step, \eqref{eqn:disadj_beuler_full} requires solving the
transposed linear system $\M^T-h\left(\frac{\partial
\f(\bu_{n+1},\btheta,t_{n+1}}{\partial \bu}\right)^T$ with respect to the
adjoint variable $\blambda_s$.

We solve both the linear systems in the forward pass and the transposed systems
in the reverse pass with a matrix-free iterative method for efficiency. The
action of the matrix or its transpose is computed by using \texttt{PyTorch}'s AD
to backpropagate $\f$. Again, only the computational graph for $\f$ needs to be
created in our approach.

\subsection{Implementation}

Drawing on state-of-the-art adjoint-capable ODE solvers (\texttt{TSAdjoint} in
\texttt{PETSc}~\cite{petsc-user-ref}), we implemented PNODE by interfacing
\texttt{PETSc} to \texttt{PyTorch} and utilizing its discrete adjoint solvers
with optimal checkpointing. As a key step, we implemented a data conversion
mechanism in \texttt{petsc4py}, the Python bindings for \texttt{PETSc}, based on
the DLPack standard.\footnote{\href{https://github.com/dmlc/dlpack}{https://github.com/dmlc/dlpack}} This enables
in-place conversion between \texttt{PETSc} vectors and \texttt{PyTorch} tensors
for both CPUs and GPUs and thus allows \texttt{PETSc} and \texttt{PyTorch} to
access the same data seamlessly. Although PNODE can be implemented with any
differentiable ODE solver such as \texttt{FATODE} \cite{Zhang2014} and
\texttt{DiffEqSensitivity.jl} \cite{DifferentialEquations.jl-2017},
\texttt{PETSc} has several favorable features that other tools lack.

\textbf{Rich set of numerical integrators}
As a widely used time-stepping library, \texttt{PETSc} offers a large collection
of time integration algorithms for solving ODEs, differential algebraic
equations, and hybrid dynamical systems \cite{abhyankar2018petsc,Zhang2017}. It
includes explicit and implicit methods, implicit-explicit methods, multirate
methods with various stability properties, and adaptive time-stepping. It also
provides a variety of advanced nonlinear/linear solvers that can be used in
PNODE. The discrete adjoint approach proposed in Section \ref{sec:disadjoint}
has been implemented for some time integrators and can be easily expanded to
others \cite{Zhang2022tsadjoint}.

\textbf{Discrete adjoint solvers with matrix-free Jacobian}
When using the adjoint solver, we compute the transposed Jacobian-vector product
through Autograd in \texttt{PyTorch} and supply it as a callback function to the
solver, instead of building the Jacobian matrix and performing matrix-vector
products, which are expensive tasks especially for dense matrices. In addition,
combining low-level AD with the high-level discrete adjoint solver guarantees
reverse accuracy, as explained in Section \ref{sec:ad}.

\textbf{Optimal checkpointing for multistage methods}
In addition to the offline binomial checkpointing algorithms
\cite{checkpointing2,Zhang2021iccs,Zhang2022TOMS}, \texttt{PETSc} supports more
sophisticated checkpointing algorithms such as online algorithms
\cite{Stumm2010} for hierarchical storage systems.

\textbf{HPC-friendly linear algebra kernels}
\texttt{PETSc} has full-fledged GPU support for efficient training of neural
ODEs, including multiprecision support (half, single, double, float128) and
extensive parallel computing that leverages CUDA-aware MPI
\cite{mills2021performanceportable}.

\section{Comparison with existing methods}

\begin{table*}
    \centering
    \small
    \caption{Comparison between different implementations of neural ODEs for $N_b$ ODE blocks. The computational complexity is measured in terms of the number of function evaluations ($\f$). }
    \label{tab:my_label}
    \begin{tabular}{c c c c c c}
    \toprule
         & NODE cont & NODE naive & ANODE & ACA & PNODE (Ours) \\
    \midrule
    Forward computation  & $N_b N_t N_s$ & $N_b N_t N_s$ & $N_b N_t N_s$ & $N_b N_t N_s$ & $N_b N_t N_s$ \\
    Reverse computation & $N_b N_t^B N_s$ & $N_b N_t N_s$ & $N_b N_t N_s$ & $N_b N_t N_s$ & $N_b N_t N_s$ \\
    Recomputation overhead & $N_b N_t^B N_s$  & $0$ & $N_b N_t N_s$ & $2N_b N_t N_s$ & $ \geq 0$ \\
    NN Backpropagation memory & $N_l$ &  $N_b N_t N_s N_l$ & $N_t N_s N_l$ & $N_s N_l$ & $ N_l$ \\
    Adjoint checkpointing memory & -- &  -- & $N_b$ & $ N_b N_t$ &  $\leq N_b (N_t\text{-}1)(N_s\text{+}1)$ \\
    Reverse-accuracy & $\color{red}\times$ & $\color{blue}\checkmark$ & $\color{blue}\checkmark$ & $\color{blue}\checkmark$ & $\color{blue}\checkmark$ \\
    Implicit time-stepping support & $\color{red}\times$  & $\color{red}\times$  & $\color{red}\times$ & $\color{red}\times$ & $\color{blue}\checkmark$ \\
    \bottomrule
    \end{tabular}
\end{table*}

A summary comparing PNODE with representative implementations of neural ODEs is
given in Table \ref{tab:my_label}. The memory and computation costs are given
for one ODE block. When multiple ODE blocks are considered, the costs need to be
multiplied by a factor of $N_b$ except for ANODE. For simplicity, we assume that
the number of reverse time steps in the continuous adjoint method is also $N_t$
and do not consider rejected time steps. Note that the rejected time steps have
no influence on the computational cost and the memory cost of PNODE because the
adjoint calculation in the reverse pass involves only accepted time steps
\cite{Zhang2021iccs}.

\textbf{NODE cont: the original implementation in \cite{NODE} with continuous
adjoint}
The vanilla neural ODE \cite{NODE} avoids recording everything by
solving the ODE backward in time to obtain the intermediate solutions needed
when solving the continuous adjoint equation. Therefore, it requires a constant
memory cost $\mathcal{O}(N_l)$ to backpropagate $\f$. The backward ODE solve
requires a recomputation cost of $\mathcal{O}(N_t^B N_s)$.

\textbf{NODE naive: a variant of the original implementation in \cite{NODE}}
This is a naive method that backpropagates the ODE solvers with low-level AD and
has the deepest computational graph, but it has no recomputational overhead.

\textbf{ANODE: a framework with discrete adjoint and checkpointing method
\cite{ANODE}}
The checkpointing method for ANODE has the same memory cost $\mathcal{O}(N_t)$
as NODE naive when a single ODE block is considered. For multiple ODE blocks,
ANODE saves only the initial states for each block and recomputes the forward
pass before the backpropagation for each block; consequently, the memory cost
for checkpointing is $\mathcal{O}(N_b)$, where $N_b$ is the number of ODE
blocks, and the backpropagation requires $\mathcal{O}(N_t N_s N_l)$ memory,
which is independent of $N_b$. Each ODE block needs to be recomputed in the
backward pass, so the total recomputation cost is $\mathcal{O}(N_t N_s)$. A
generalization of ANODE has been implemented in a Julia library
\cite{rackauckas2020universal}.

\textbf{ACA: the adaptive checkpoint adjoint (ACA) method \cite{zhuang2020}}
ACA is similar to ANODE but uses a slightly different checkpointing strategy.
ACA checkpoints the state at each time step, thus consuming $\mathcal{O}(N_t)$
memory for checkpointing. To save memory, ACA deletes redundant computational
graphs for rejected time steps when searching for the optimal step size.
Although it is shown in \cite{zhuang2020} that ACA uses the continuous adjoint
approach, the actual implementation of ACA applies backpropagation with
low-level AD to each time step, so it requires $\mathcal{O}(N_s N_l)$ memory. In
the backward pass, ACA first performs an additional forward pass to save the
checkpoints and then recomputes each time step to generate the local
computational graph, resulting in a total recomputation cost of
$\mathcal{O}(2N_t N_s)$.

\section{Experimental Results} \label{sec:exp}

In this section we test the performance of PNODE and compare it with existing
neural ODE methods on two distinct benchmark tasks: image classification and
continuous normalizing flows (CNF). We then demonstrate the application of PNODE
to learning stiff dynamics. All the experiments are conducted on an NVIDIA Tesla
V100 GPU with a memory capacity of 32 GB. When testing PNODE, unless specified
otherwise, we assume $N_c \geq N_t$ and checkpoint all the intermediate
solutions and stage values for the best speed. Note that this approach leads to
no recomputation, thus giving the best speed but also the worst-case memory cost
for PNODE.

Existing frameworks may support only a subset of the time integration schemes
used in our experiments. For a comprehensive and fair evaluation and comparison,
we have added the same time integration schemes in all frameworks if they were
not originally implemented. In the benchmark tests, we focused on fixed step
schemes instead of adaptive step schemes because error estimation and adaptive
strategies can be vastly different across different frameworks, making a fair
comparison difficult, especially when investigating memory consumption, which is
sensitive to the number of time steps. The results for fixed step schemes can
reflect the fundamental differences among the neural ODE methods and indicate
the expected performance of these methods for adaptive step schemes.

\subsection{Image classification}

\begin{figure}
  \centering
  \includegraphics[width=0.7\linewidth]{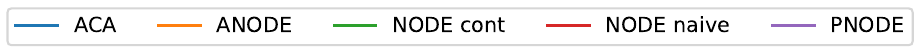}
  \includegraphics[width=0.34\linewidth]{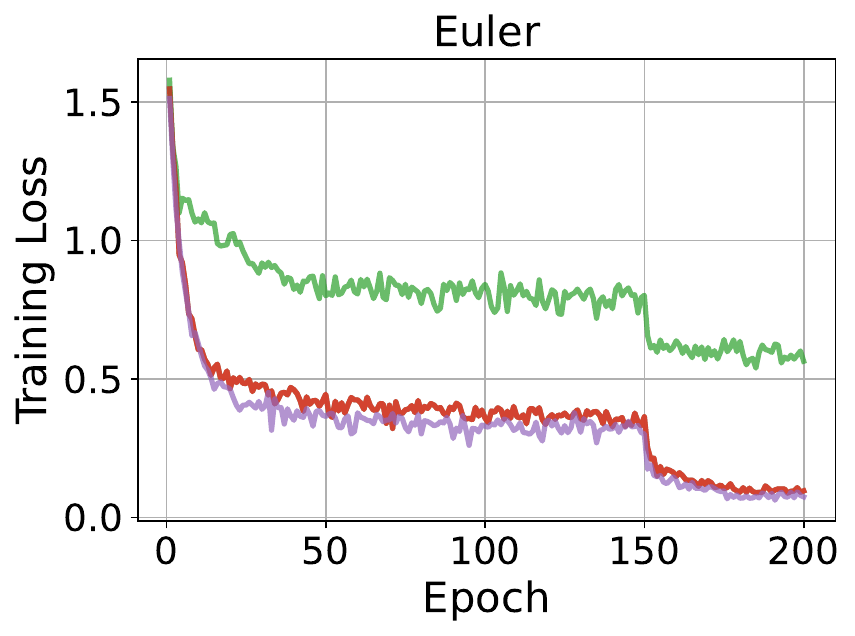}
  \includegraphics[width=0.34\linewidth]{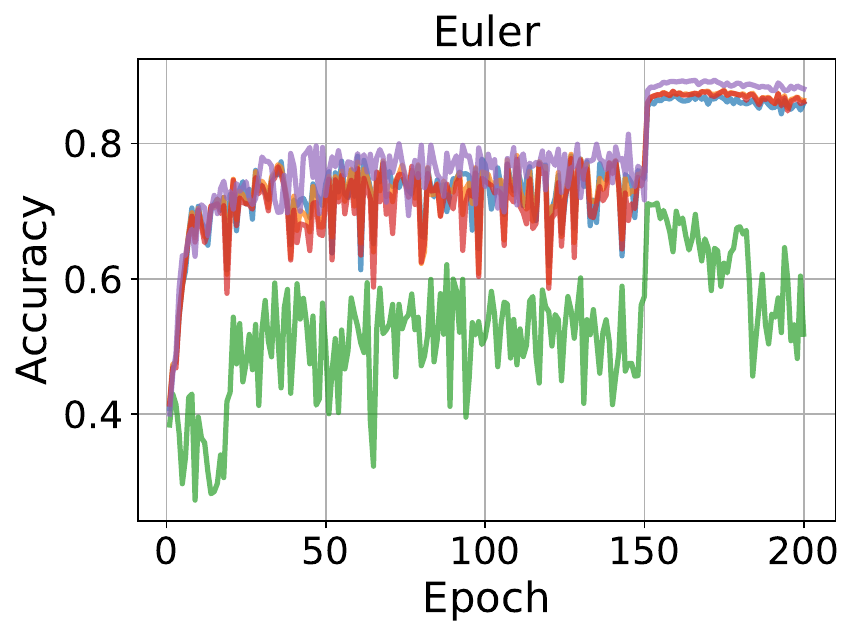}
  \includegraphics[width=0.34\linewidth]{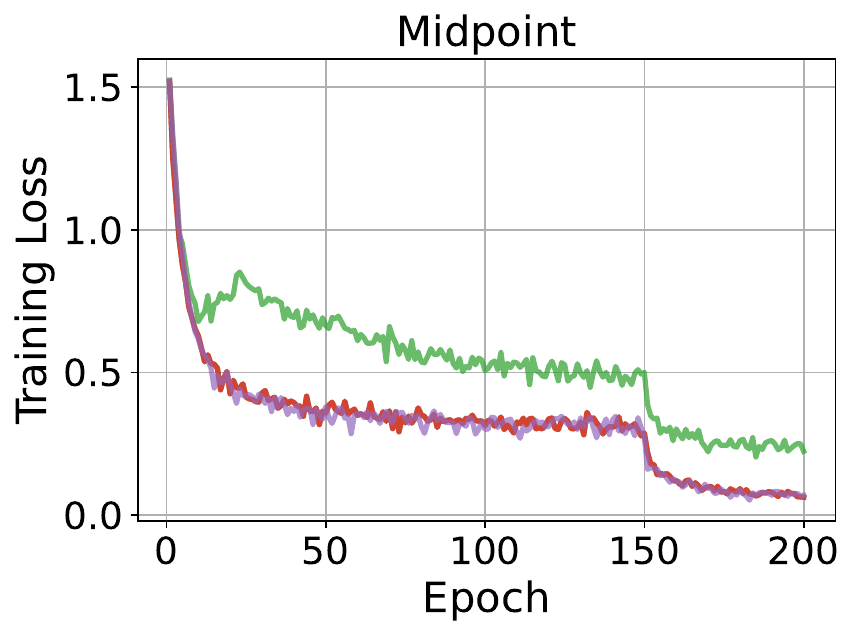}
  \includegraphics[width=0.34\linewidth]{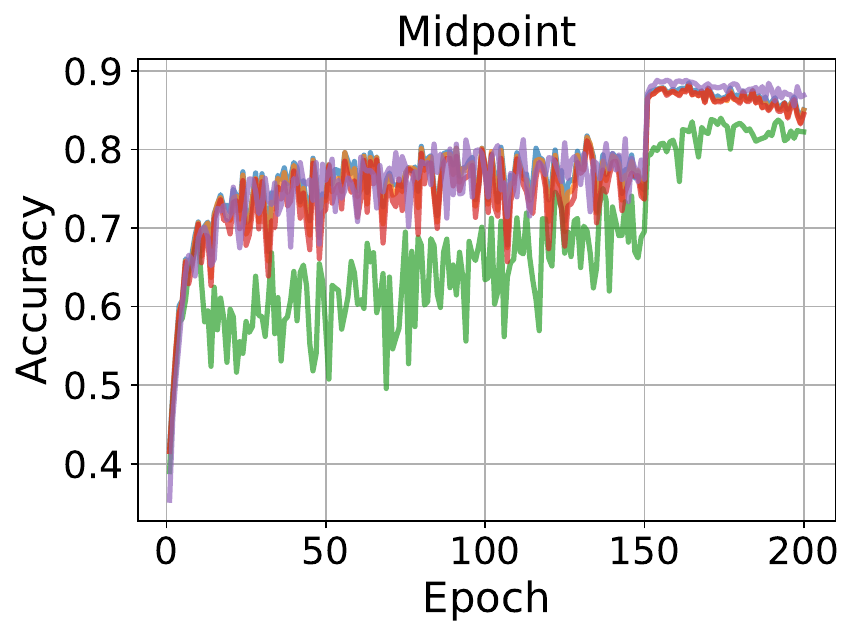}
  \includegraphics[width=0.34\linewidth]{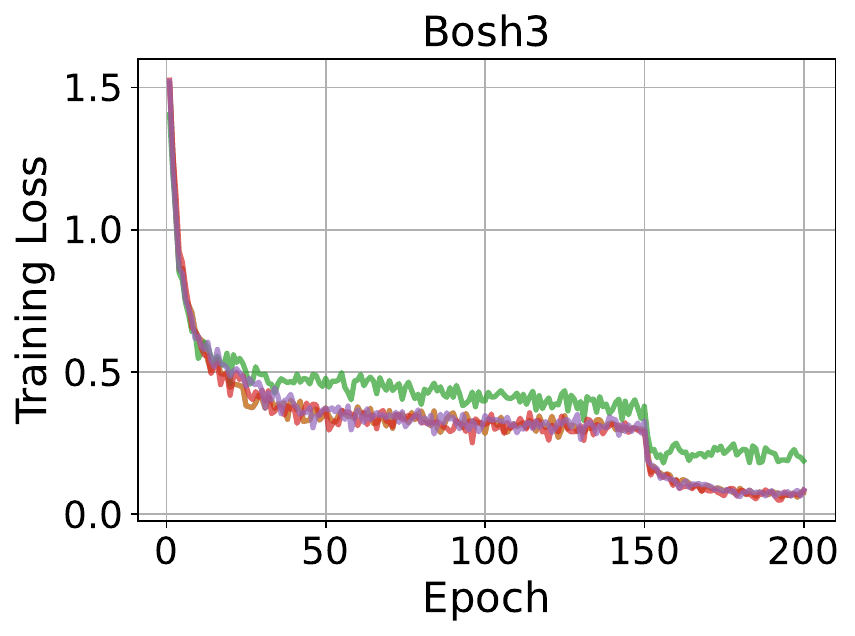}
  \includegraphics[width=0.34\linewidth]{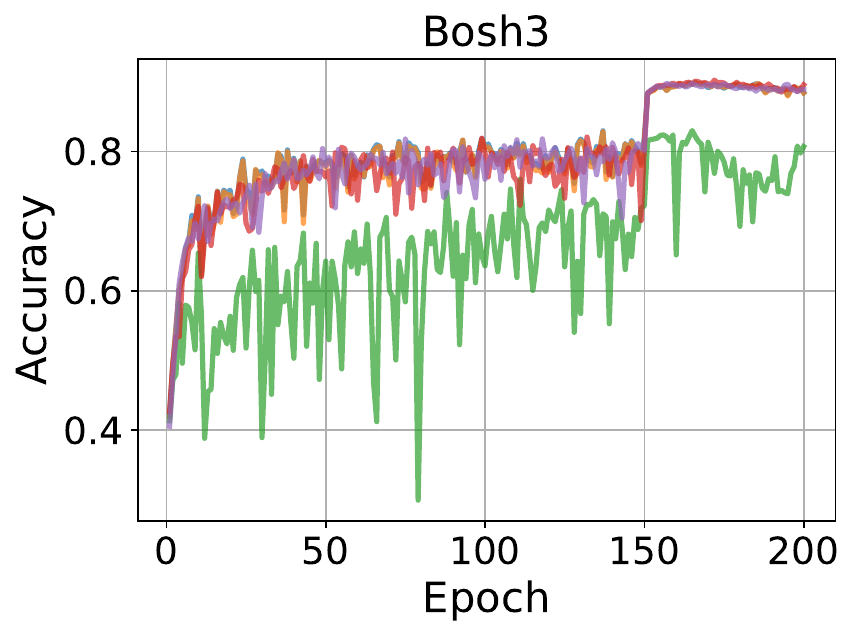}
  \includegraphics[width=0.34\linewidth]{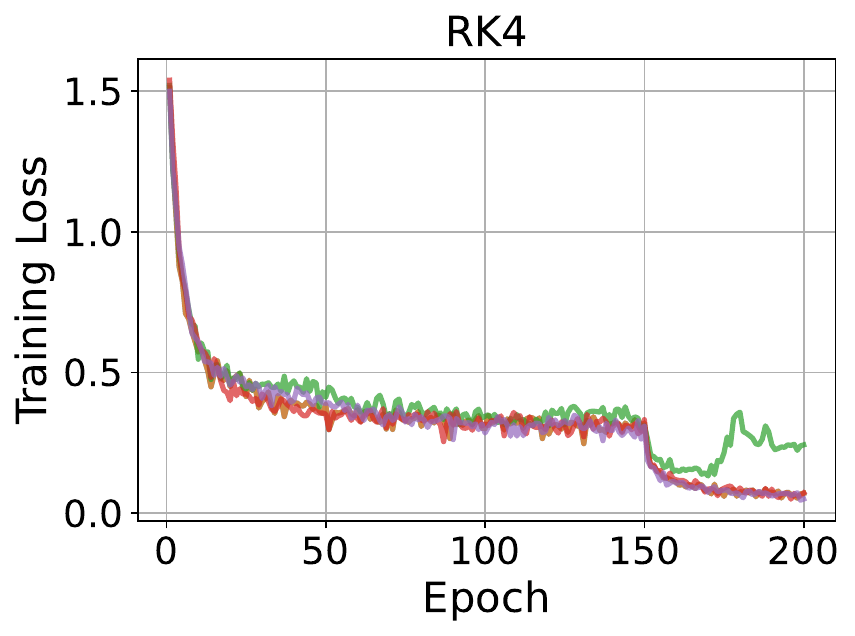}
  \includegraphics[width=0.34\linewidth]{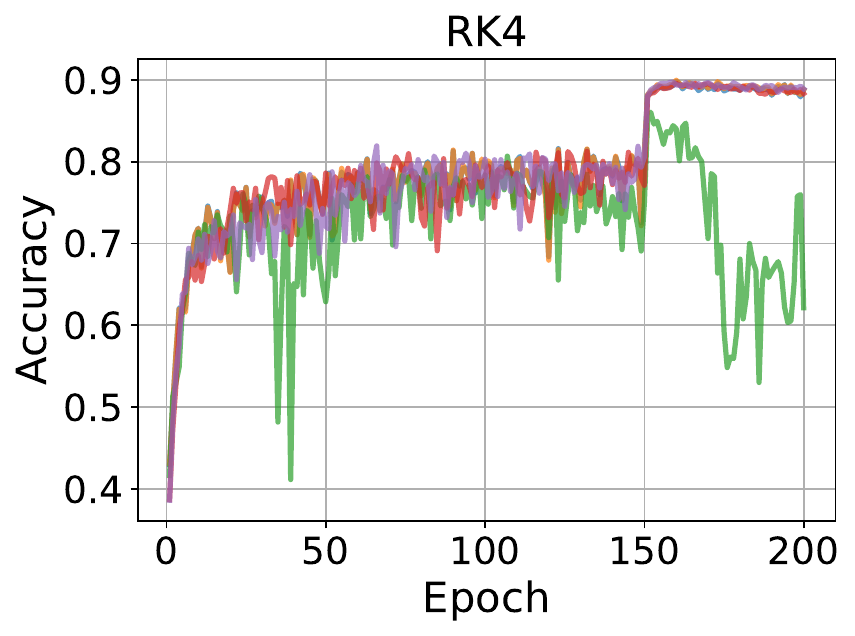}
  \includegraphics[width=0.34\linewidth]{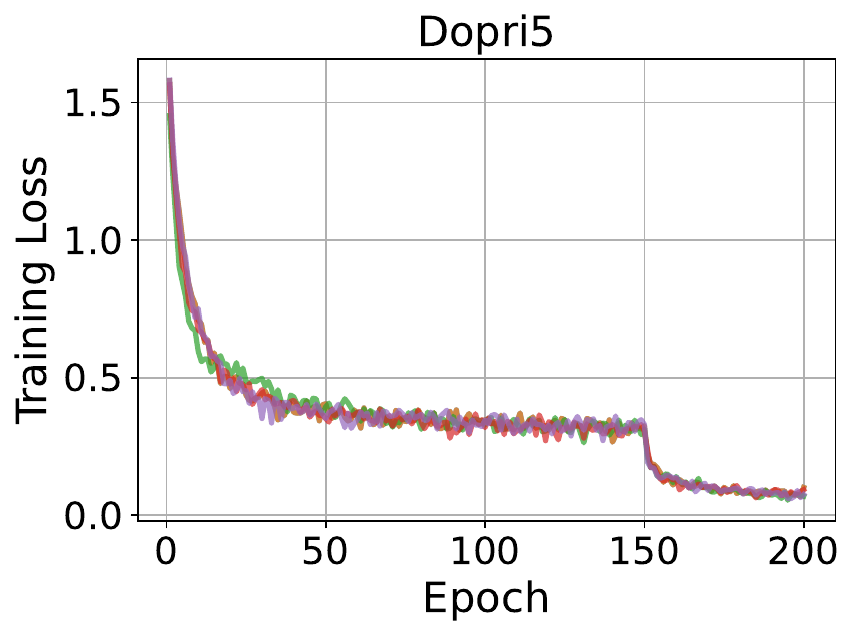}
  \includegraphics[width=0.34\linewidth]{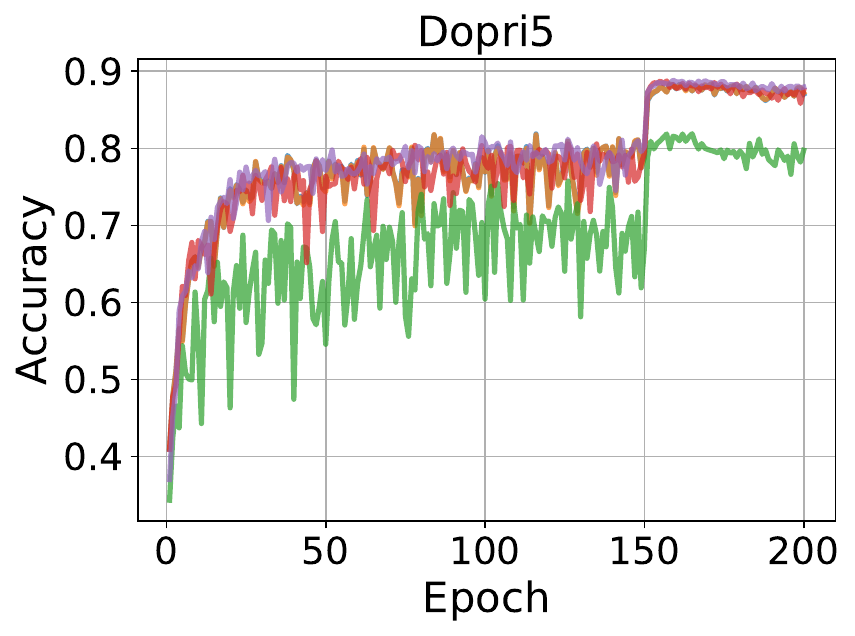}
  \caption{Training and testing performance of SqueezeNext on CIFAR10 using various schemes with one time step. The inaccuracy in the gradient calculated via continuous adjoint causes a significant gap in performance between discrete adjoint and continuous adjoint.
  }
  \label{fig:cifar10}
\end{figure}

Chen et al. \cite{NODE} showed that replacing residual blocks with ODE blocks
yields good accuracy and parameter efficiency on the MNIST dataset from
\cite{MNIST}. However, the effect of the accuracy of the gradients cannot be
determined easily with this simple dataset. Here we experiment on the more
complex CIFAR-10 dataset from \cite{CIFAR} using a SqueezeNext network
introduced by \cite{SqueezeNext}, where every nontransition block is replaced
with a neural ODE block. We use $4$ ODE blocks of different dimensions with
$199,800$ trainable parameters in total.

\paragraph{Accuracy} We train the same neural ODE structure using one time step
while varying the time integration method. The results are displayed in Figure
\ref{fig:cifar10}. The ReLU activation in this model results in irreversible
dynamics and inaccurate gradient calculation for the vanilla neural ODE
\cite{NODE}, leading to divergent training with the Euler method and the RK4
method and to suboptimal accuracy. PNODE and other neural ODE methods converge
to higher accuracy because of the reverse accuracy guaranteed by the discrete
adjoint method and automatic differentiation. With the low-accuracy methods
(Euler and Midpoint), PNODE achieves the highest accuracy among all the methods
tested.

\begin{figure}[htbp]
  \centering
  \includegraphics[width=0.6\linewidth]{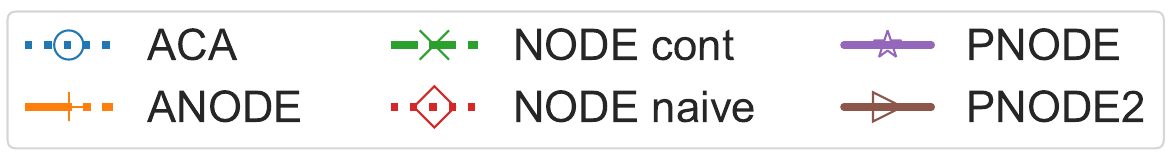}
  \includegraphics[width=0.6\linewidth]{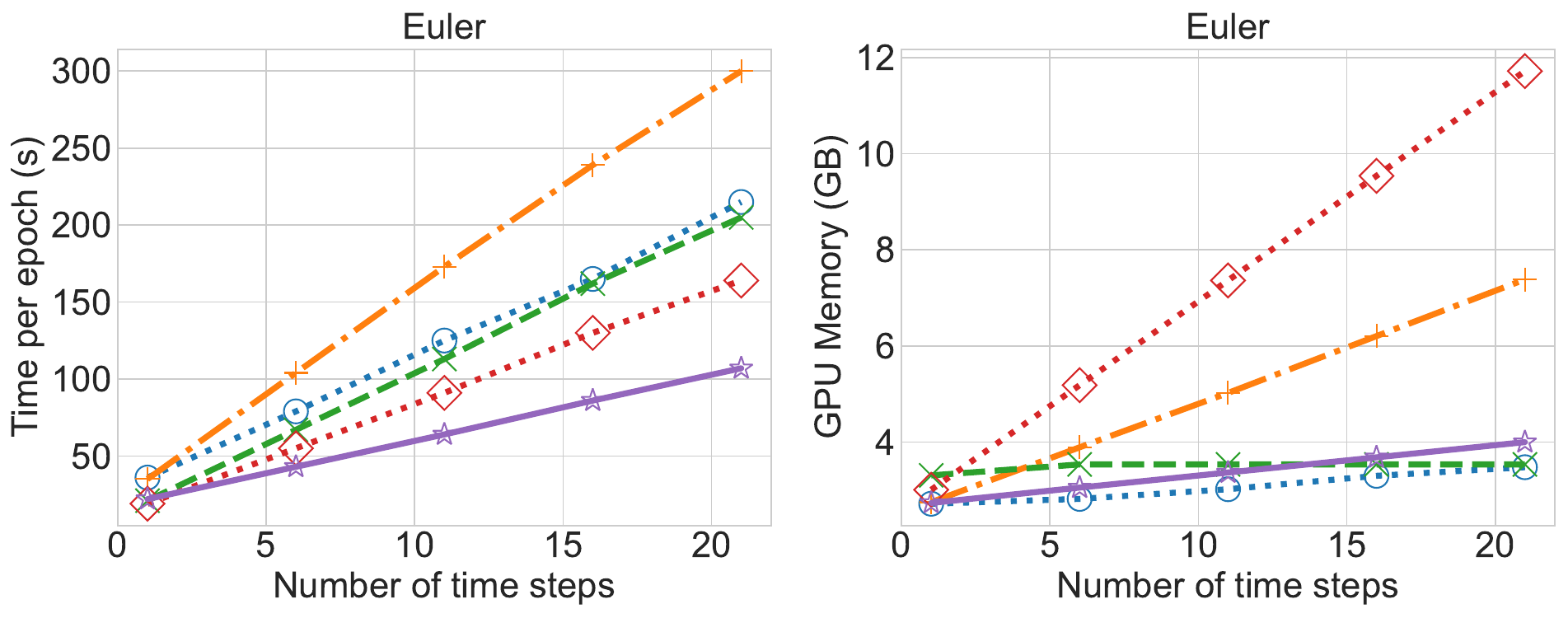}
  \includegraphics[width=0.6\linewidth]{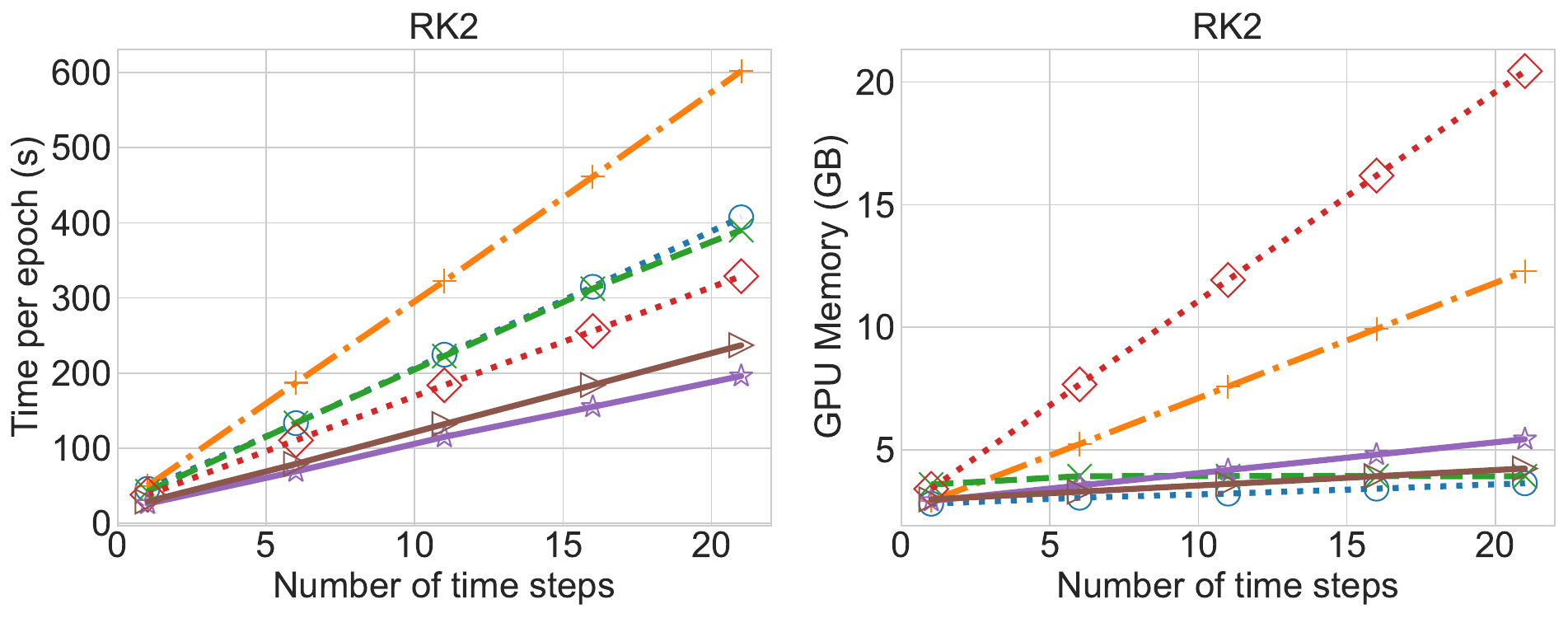}
  \includegraphics[width=0.6\linewidth]{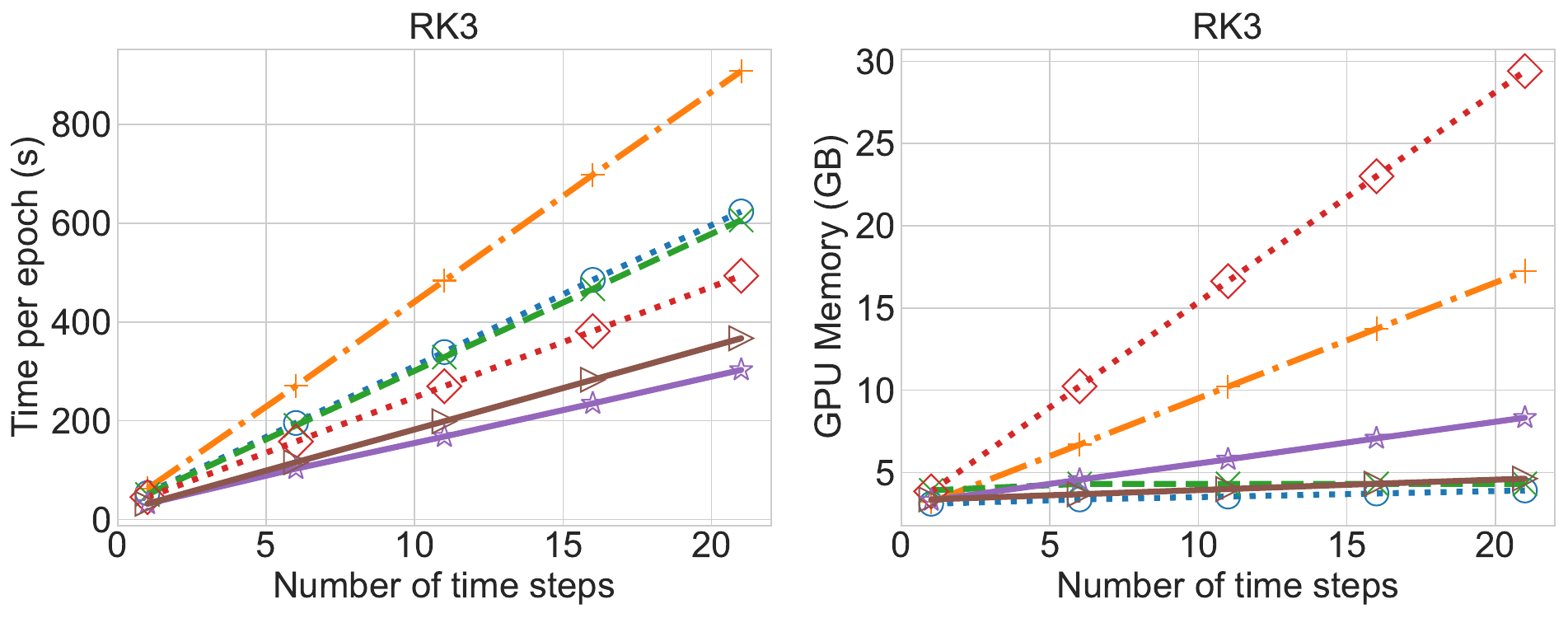}
  \includegraphics[width=0.6\linewidth]{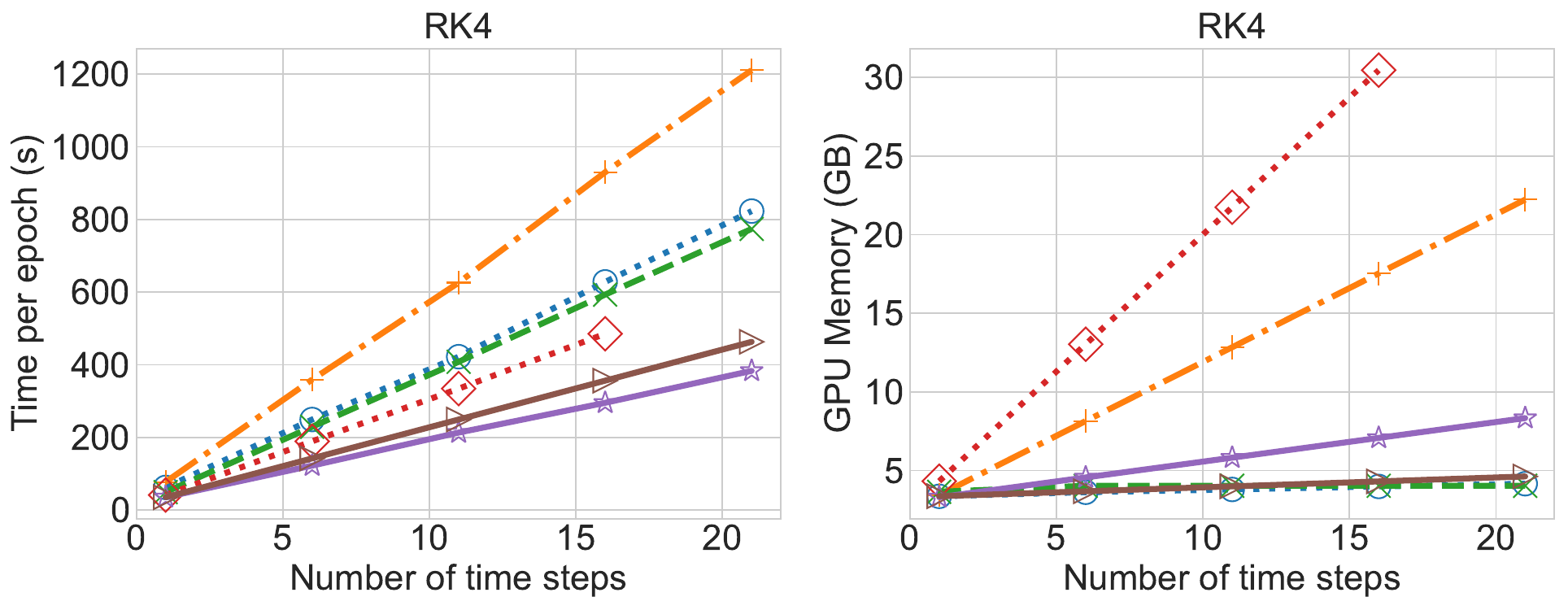}
  \includegraphics[width=0.6\linewidth]{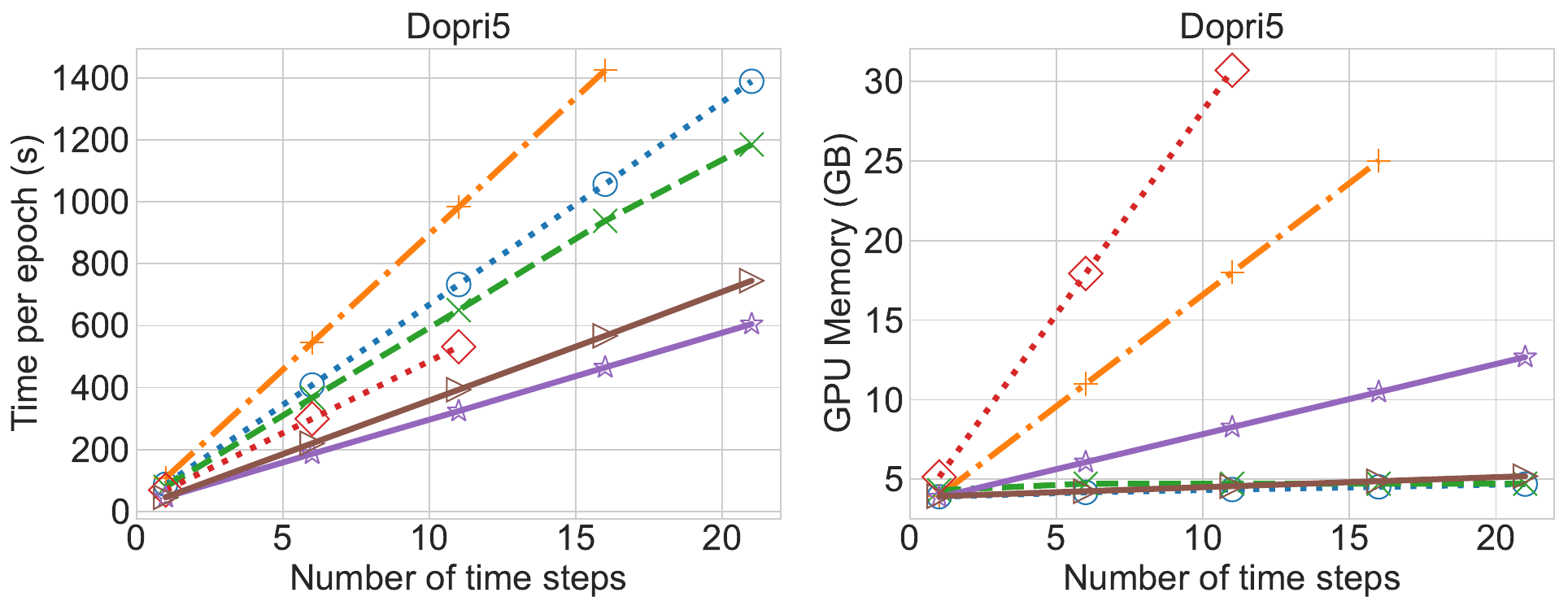}
  \caption{GPU memory usage and time per epoch of different implementations as functions of the number of time steps ($N_t$) for various schemes.
  }
  \label{fig:cifar10-efficiency}
\end{figure}

\paragraph{Memory/time efficiency} We perform a systematic comparison of the
efficiency and memory cost for different methods by varying $N_t$. In all the
experiments, we measure the total amount of consumed GPU memory. For PNODE, we
include an additional variant (denoted by PNODE2) that saves only the ODE
solution at each time step. This leads to $N_t\text{-}1$ recomputations in the
reverse pass and a memory cost of $N_b(N_t\text{-}1)$ for adjoint checkpointing.
Thus PNODE2 has almost the same total memory cost as ACA has. As shown in Figure
\ref{fig:cifar10-efficiency}, PNODE significantly outperforms all the other
methods in terms of per-epoch training time. When using Dopri5, PNODE is three
times faster than ANODE, $2.3$ times faster than ACA, and two times faster than
the vanilla neural ODE. Similar speedups can be observed for other time
integration methods. Among all reverse-accurate methods, PNODE has the slowest
memory growth as the number of time steps increases. For example, using Dopri5
with $11$ time steps, the memory consumption of PNODE is approximately $71\%$
less than NODE naive and $55\%$ less than ANODE. Compared with PNODE, PNODE2
reduces the memory cost by up to $42\%$ with a slight increase in training time.
PNODE2 has a similar memory cost but much faster training speed than ACA, which
agrees with the theoretical analysis shown in Table \ref{tab:my_label}. Note
that the CUDA runtime allocates $\sim 0.4$ GB memory for PNODE, and it is
included in all the PNODE results presented in this paper. This overhead is
inevitable for loading any library that contains CUDA kernels.

\subsection{Continuous normalizing flow for density estimation}

We select three datasets---POWER, MINIBOONE, and BSDS3000---that are
respectively 6-, 43-, and 63-dimensional tabular datasets commonly used in CNF
\cite{MAF}. The FFJORD \cite{grathwohl2018scalable} approach is used to
transform a multivariate Gaussian distribution to the target distributions for
all three datasets.

We adopt the tuned architecture and hyperparameters (e.g., learning rates,
number of hidden layers, and number of flow steps) from
\cite{grathwohl2018scalable}. We vary the number of steps for different time
integration methods only for stability considerations. On all three datasets,
all the methods that converge yield comparable testing losses, but all the
reverse-accurate methods converge much faster than NODE cont because of more
accurate gradient estimates. On BSDS300, NODE cont failed to converge after 14
days of training and was terminated prematurely. Similar observations were
reported in \cite{onken2020discretizeoptimize}.

The performance statistics including the number of function evaluations per
iteration, the training time per iteration, and the maximum GPU memory usage are
shown in \Cref{tab:CNF1,tab:CNF2,tab:CNF3,tab:CNF4,tab:CNF5}. NFE-F and NFE-B
are the number of function evaluations in the forward pass and backward pass,
respectively. NFE-B for PNODE and NODE cont reflects the cost of the transposed
Jacobian-vector products, while NFE-B for other methods reflects the cost for
recomputing the time steps in the backward pass. The observed results agree with
our theoretical analysis in Table \ref{tab:my_label}.

Excluding the $0.4$ GB constant overhead, PNODE has the lowest memory
consumption among all reverse-accurate methods, and it consistently outperforms
ACA and ANODE in terms of training time for all three datasets. When using the
Dopri5 scheme that is commonly used in neural ODEs, PNODE is $28.7\%$ faster and
consumes $68.2\%$ less memory than does ACA for BSDS300. The advantage becomes
more evident for high-order schemes that have more stages. The reason is that
the computational graph for automatic differentiation grows deeper as the number
of stages increases.

NODE naive and ANODE run out of GPU memory for BSDS300. But when memory is
sufficient, NODE naive is the fastest, as expected. For MINIBOONE, the
difference in memory consumption is marginal for all methods because of a number
of factors such as small batch size ($1000$), small number of steps ($4$), and
small neural network.

\begin{table}[ht]
  \caption{Performance statistics for the Euler scheme. Missing values are due to out-of-memory errors.}
  \label{tab:CNF2}
  \centering
  \begin{tabular}{llccccc}
    \toprule
    \multirow{2}{*}{Dataset } & \multirow{2}{*}{Framework} & \multirow{2}{*}{ \begin{minipage}{0.6in}Integration \\ method\end{minipage} } & \multirow{2}{*}{NFE-F} & \multirow{2}{*}{NFE-B} & \multirow{2}{*}{  \begin{minipage}{0.65in} Time per \\iteration (s) \end{minipage} } & \multirow{2}{*}{\begin{minipage}{0.65in} GPU Mem \\ (GB) \end{minipage}} \\ \\
    \midrule
    \multirow{3}{*}{POWER} & NODE naive & \multirow{5}{*}{ \begin{minipage}{0.6in} Euler, $N_t$=50\end{minipage} }& 250 & 0 & 0.760 & 11.648 \\
    & NODE cont & & 250 & 250 & 1.101 & 1.680 \\
    & ANODE & & 250 & 250 & 1.297 & 3.624 \\
    & ACA & & 250 & 505 & 1.382 & 1.735 \\
    & PNODE & & 250 & 250 & \textbf{1.117} & \textbf{2.104} \\
    \midrule
    \multirow{3}{*}{MINIBOONE} & NODE naive & \multirow{5}{*}{ \begin{minipage}{0.6in} Euler, $N_t$=20\end{minipage} } & 20 & 0 & 0.071 & 2.481 \\
    & NODE cont & & 20 & 20 & 0.094 & 1.716 \\
    & ANODE & & 20 & 20 & 0.108 & 2.500 \\
    & ACA & & 20 & 41 & 0.102 & 1.762 \\
    & PNODE & & 20 & 20 & \textbf{0.099} & \textbf{2.085} \\
    \midrule
    \multirow{3}{*}{BSDS300} & NODE naive & \multirow{5}{*}{ \begin{minipage}{0.6in} Euler, $N_t$=100\end{minipage} } & -- & -- & -- & -- \\
    & NODE cont & & 200 & 200 & 13.409 & 3.826 \\
    & ANODE & & -- & -- & -- & -- \\
    & ACA & & 200 & 402 & 17.191 & 5.564 \\
    & PNODE & & 200 & 200 & \textbf{13.541} & \textbf{4.920} \\
    \bottomrule
  \end{tabular}
\end{table}
\begin{table}[ht]
  \caption{Performance statistics for the Midpoint scheme.}
  \label{tab:CNF3}
  \centering
  \begin{tabular}{llccccc}
    \toprule
    \multirow{2}{*}{Dataset } & \multirow{2}{*}{Framework} & \multirow{2}{*}{ \begin{minipage}{0.6in}Integration \\ method\end{minipage} } & \multirow{2}{*}{NFE-F} & \multirow{2}{*}{NFE-B} & \multirow{2}{*}{  \begin{minipage}{0.65in} Time per\\ iteration (s) \end{minipage} } & \multirow{2}{*}{\begin{minipage}{0.65in} GPU Mem \\ (GB) \end{minipage}} \\ \\
    \midrule
    \multirow{3}{*}{POWER} & NODE naive & \multirow{5}{*}{ \begin{minipage}{0.6in} Midpoint, $N_t$=40\end{minipage} }& 400 & 0 & 1.368 & 17.659 \\
    & NODE cont & & 400 & 400 & 1.780 & 1.680 \\
    & ANODE & & 400 & 400 & 1.975 & 4.822 \\
    & ACA & & 400 & 805 & 2.109 & 1.819 \\
    & PNODE & & 400 & 400 & \textbf{1.883} & \textbf{2.152} \\
    \midrule
    \multirow{3}{*}{MINIBOONE} & NODE naive & \multirow{5}{*}{ \begin{minipage}{0.6in} Midpoint, $N_t$=16\end{minipage} } & 32 & 0 & 0.104 & 2.991 \\
    & NODE cont & & 32 & 32 & 0.134 & 1.737 \\
    & ANODE & & 32 & 32 & 0.167 & 3.012 \\
    & ACA & & 32 & 65 & 0.145 & 1.846 \\
    & PNODE & & 32 & 32 & \textbf{0.137} & \textbf{2.087} \\
    \midrule
    \multirow{3}{*}{BSDS300} & NODE naive & \multirow{5}{*}{ \begin{minipage}{0.6in} Midpoint, $N_t$=80\end{minipage} } & -- & -- & -- & -- \\
    & NODE cont & & 320 & 320 & 21.404 & 3.826 \\
    & ANODE & & -- & -- & -- & -- \\
    & ACA & & 320 & 642 & 24.977 & 7.525 \\
    & PNODE & & 320 & 320 & \textbf{21.651} & \textbf{5.388} \\
    \bottomrule
  \end{tabular}
\end{table}
\begin{table}[ht]
  \caption{Performance statistics for the Bosh3 scheme.}
  \label{tab:CNF4}
  \centering
  \begin{tabular}{llccccc}
    \toprule
    \multirow{2}{*}{Dataset } & \multirow{2}{*}{Framework} & \multirow{2}{*}{ \begin{minipage}{0.6in}Integration \\ method\end{minipage} } & \multirow{2}{*}{NFE-F} & \multirow{2}{*}{NFE-B} & \multirow{2}{*}{  \begin{minipage}{0.65in} Time per\\ iteration (s) \end{minipage} } & \multirow{2}{*}{\begin{minipage}{0.65in} GPU Mem (GB) \end{minipage}} \\ \\
    \midrule
    \multirow{3}{*}{POWER} & NODE naive & \multirow{5}{*}{ \begin{minipage}{0.6in} Bosh3, $N_t$=30\end{minipage} }& 465 & 0 & 1.532 & 20.282 \\
    & NODE cont & & 465 & 465 & 3.058 & 6.555 \\
    & ANODE & & 450 & 450 & 2.593 & 10.106 \\
    & ACA & & 450 & 905 & 2.632 & 6.771 \\
    & PNODE & & 455 & 450 & \textbf{2.084} & \textbf{2.217} \\
    \midrule
    \multirow{3}{*}{MINIBOONE} & NODE naive & \multirow{5}{*}{ \begin{minipage}{0.6in} Bosh3, $N_t$=12\end{minipage} } & 32 & 0 & 0.1041 & 2.991 \\
    & NODE cont & & 39 & 39 & 0.164 & 1.737 \\
    & ANODE & & 36 & 36 & 0.187 & 3.182 \\
    & ACA & & 36 & 73 & 0.219 & 1.930 \\
    & PNODE & & 37 & 36 & \textbf{0.153} & \textbf{2.091} \\
    \midrule
    \multirow{3}{*}{BSDS300} & NODE naive & \multirow{5}{*}{ \begin{minipage}{0.6in} Bosh3, $N_t$=60\end{minipage} } & -- & -- & -- & -- \\
    & NODE cont & & 366 & 366 & 22.905 & 3.878 \\
    & ANODE & & -- & -- & -- & -- \\
    & ACA & & 360 & 722 & 31.001 & 9.821 \\
    & PNODE & & 362 & 360 & \textbf{24.439} & \textbf{6.145} \\
    \bottomrule
  \end{tabular}
\end{table}
\begin{table}[ht]
  \caption{Performance statistics for the RK4 scheme.}
  \label{tab:CNF5}
  \centering
  \begin{tabular}{llccccc}
    \toprule
    \multirow{2}{*}{Dataset } & \multirow{2}{*}{Framework} & \multirow{2}{*}{ \begin{minipage}{0.6in}Integration \\ method\end{minipage} } & \multirow{2}{*}{NFE-F} & \multirow{2}{*}{NFE-B} & \multirow{2}{*}{  \begin{minipage}{0.65in} Time per\\ iteration (s) \end{minipage} } & \multirow{2}{*}{\begin{minipage}{0.65in} GPU Mem (GB) \end{minipage}} \\ \\
    \midrule
    \multirow{3}{*}{POWER} & NODE naive & \multirow{5}{*}{ \begin{minipage}{0.6in} RK4, $N_t$=20\end{minipage} }& 400 & 0 & 1.615 & 22.529 \\
    & NODE cont & & 400 & 400 & 2.291 & 6.553 \\
    & ANODE & & 400 & 400 & 2.345 & 9.697 \\
    & ACA & & 400 & 805 & 2.365 & 6.834 \\
    & PNODE & & 400 & 400 & \textbf{1.765} & \textbf{2.156} \\
    \midrule
    \multirow{3}{*}{MINIBOONE} & NODE naive & \multirow{5}{*}{ \begin{minipage}{0.6in} RK4, $N_t$=8\end{minipage} } & 32 & 0 & 0.110 & 2.991 \\
    & NODE cont & & 32 & 32 & 0.131 & 1.737 \\
    & ANODE & & 32 & 32 & 0.168 & 3.012 \\
    & ACA & & 32 & 65 & 0.200 & 2.016 \\
    & PNODE & & 32 & 32 & \textbf{0.135} & \textbf{2.089} \\
    \midrule
    \multirow{3}{*}{BSDS300} & NODE naive & \multirow{5}{*}{ \begin{minipage}{0.6in} RK4, $N_t$=40\end{minipage} } & -- & -- & -- & -- \\
    & NODE cont & & 320 & 320 & 21.097 & 3.878 \\
    & ANODE & & -- & -- & -- & -- \\
    & ACA & & 320 & 642 & 27.035 & 12.053 \\
    & PNODE & & 320 & 320 & \textbf{21.681} & \textbf{5.422} \\
    \bottomrule
  \end{tabular}
\end{table}
\begin{table}[ht]
    \caption{Performance statistics for the Dopri5 scheme.}
    \label{tab:CNF1}
    \centering
    \begin{tabular}{llccccc}
      \toprule
      \multirow{2}{*}{Dataset } & \multirow{2}{*}{Framework} & \multirow{2}{*}{ \begin{minipage}{0.6in}Integration \\ method\end{minipage} } & \multirow{2}{*}{NFE-F} & \multirow{2}{*}{NFE-B} & \multirow{2}{*}{  \begin{minipage}{0.65in} Time per\\ iteration (s) \end{minipage} } & \multirow{2}{*}{\begin{minipage}{0.65in} GPU Mem (GB) \end{minipage}} \\ \\
      \midrule
      \multirow{3}{*}{POWER} & NODE naive & \multirow{5}{*}{ \begin{minipage}{0.6in} Dopri5, $N_t$=10\end{minipage} }& 300 & 0 & 0.976 & 13.653 \\
      & NODE cont & & 300 & 300 & 1.357 & 1.687 \\
      & ANODE & & 300 & 300 & 1.524 & 4.035 \\
      & ACA & & 300 & 605 & 1.816 & 2.131 \\
      & PNODE & & 305 & 300 & \textbf{1.329} & \textbf{2.150} \\
      \midrule
      \multirow{3}{*}{MINIBOONE} & NODE naive & \multirow{5}{*}{ \begin{minipage}{0.6in} Dopri5, $N_t$=4\end{minipage} } & 24 & 0 & 0.074 & 2.653 \\
      & NODE cont & & 24 & 24 & 0.120 & 1.758 \\
      & ANODE & & 24 & 24 & 0.131 & 2.672 \\
      & ACA & & 24 & 49 & 0.143 & 2.184 \\
      & PNODE & & 25 & 24 & \textbf{0.105} & \textbf{2.089} \\
      \midrule
      \multirow{3}{*}{BSDS300} & NODE naive & \multirow{5}{*}{ \begin{minipage}{0.6in} Dopri5, $N_t$=20\end{minipage} } & -- & -- & -- & -- \\
      & NODE cont & & 240 & 240 & 16.175 & 3.983 \\
      & ANODE & & -- & -- & -- & -- \\
      & ACA & & 240 & 482 & 21.003 & 16.667 \\
      & PNODE & & 242 & 240 & \textbf{16.314} & \textbf{5.304} \\
      \bottomrule
    \end{tabular}
  \end{table}

\subsection{Learning stiff dynamics}

Stiff dynamical systems are characterized by widely separated time scales, which
pose computational difficulties for explicit time integration methods. Several
researchers   \cite{Ghosh2020_NIPS,Kim2021,Liang2022} have recognized that
learning stiff dynamics from time-series data is also challenging for
data-driven approaches such as neural ODEs. The reasons are twofold. First, the
computational difficulties in solving stiff ODEs remain in neural ODEs. Second,
stiffness could lead to ill-conditioned gradients in classical neural ODEs. Kim
et al.~\cite{Kim2021} proposed applying scaling to the differential equations
and loss functions to mitigate stiffness. In this section we apply \emph{feature
scaling} to the input data and then use an implicit method for training the
neural ODEs. We demonstrate the success of our approach with a well-known stiff
chemical reaction system and illuminate the benefits of using implicit solvers,
which are equipped with discrete adjoint capabilities and efficient nonlinear
and linear solvers, uniquely enabled through PNODE.

The system we consider is governed by Robertson's equations:
\begin{equation}
\begin{aligned}
    \frac{d u_{1}}{d t} &= -k_{1} u_{1}+k_{3} u_{2} u_{3}, \\
    \frac{d u_{2}}{d t} &= k_{1} u_{1}-k_{2} u_{2}^{2}-k_{3} u_{2} u_{3}, \\
    \frac{d u_{3}}{d t} &= k_{2} u_{2}^{2} ,
\end{aligned}
\label{eq:rober}
\end{equation}
where $u_1, u_2, u_3$ are the concentrations of three species and $k_1=0.04,
k_2=3\times10^7, k_3=10^4$ are reaction rate constants. We train a neural ODE in
form \ref{eq:ode} with the right-hand side function approximated by a NN. The
loss function is defined as the discrepancy between the observed data and the
prediction by the ML model
\begin{equation}
  \mathcal{L} = \text{MAE}(u^{\text{ob}}(t), u^{\text{pred}}(t)) ,
\end{equation}
where $\text{MAE}$ is the mean absolute error. The NN has five
hidden layers with an activation function of GELU as used in \cite{Kim2021}.

The training data are generated by solving \eqref{eq:rober} with the initial
conditions of $[u_1, u_2, u_3] = [1, 0, 0]$ over a time span of $[10^{-5}, 100]$
and sampling $40$ data points equally spaced in a logarithm scale. We train the
neural ODE for $10,000$ epochs on a GPU using the AdamW optimizer with an
initial learning rate of $0.005$.

\subsubsection{Data preprocessing}

\begin{figure*}
  \centering
  \subfigure[Implicit method]{
  \includegraphics[width=0.31\textwidth]{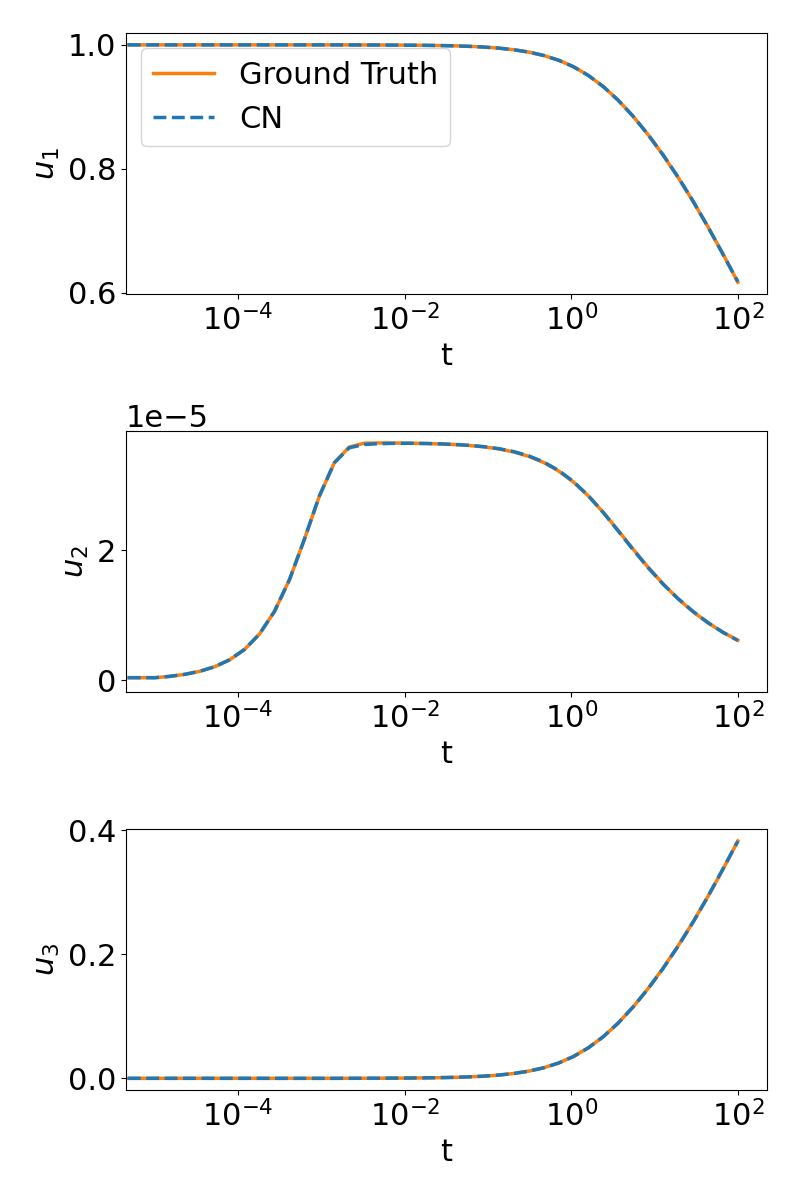}
  }
  \subfigure[Explicit method]{
  \includegraphics[width=0.31\textwidth]{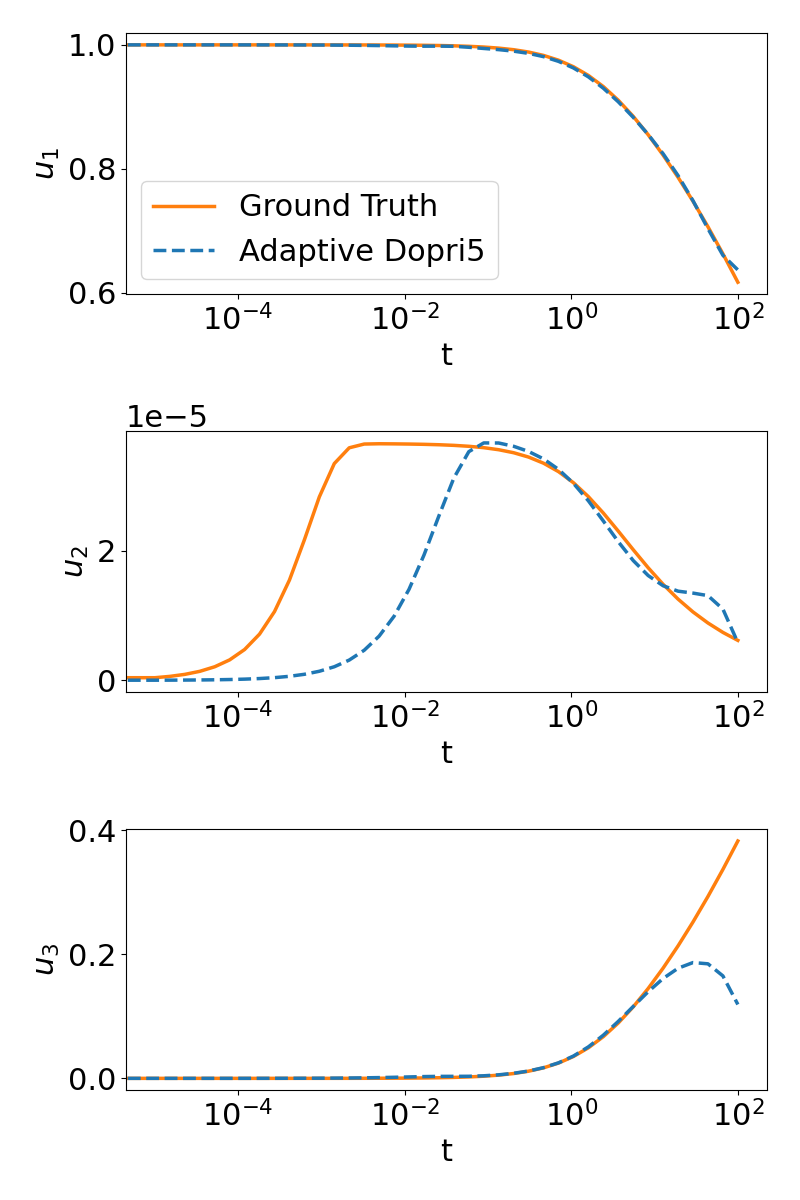}
  }
  \subfigure[Implicit method (no scaling)]{
  \includegraphics[width=0.31\textwidth]{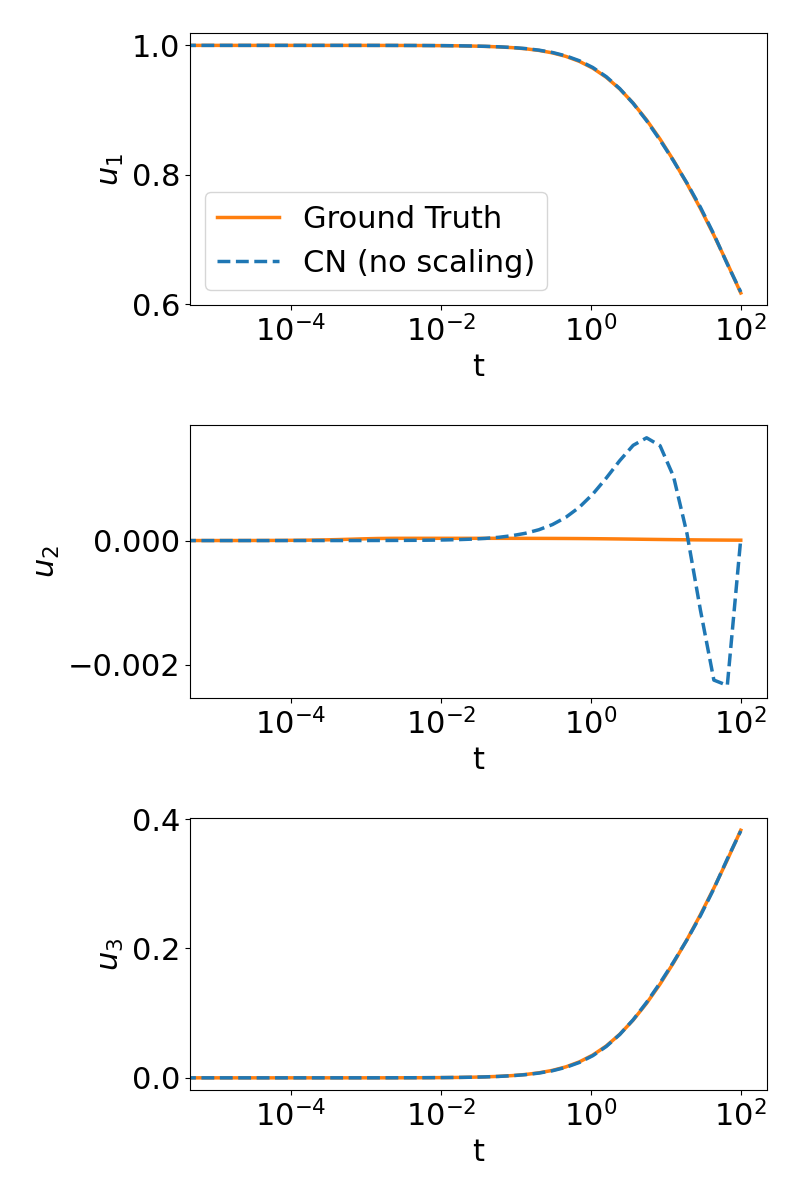}
  }
  \caption{Ground truth of Robertson's equations and the predictions of the neural ODE models with different training methods.}
  \label{fig:rober-accuracy}
\end{figure*}

Since the three species have widely varying concentrations (5 orders of
magnitude), their contributions to the loss function are not proportionate, thus
leading to slow convergence in training. Figure \ref{fig:rober-accuracy}(c)
shows the results learned by using the raw data after $10,000$ epochs' training.
The predictions for the first and third species match the ground truth well;
however, the second species shows a noticeable discrepancy. To remedy this
issue, we apply the standard min-max normalization
\begin{equation}
  u^{\prime}=\frac{u-\min (u)}{\max (u)-\min (u)}
\end{equation}
to scale the input data to the range $[0,1]$.

\subsubsection{Implicit versus explicit}

\begin{figure}
    \centering
    \subfigure[CN]{
    \includegraphics[width=0.38\linewidth]{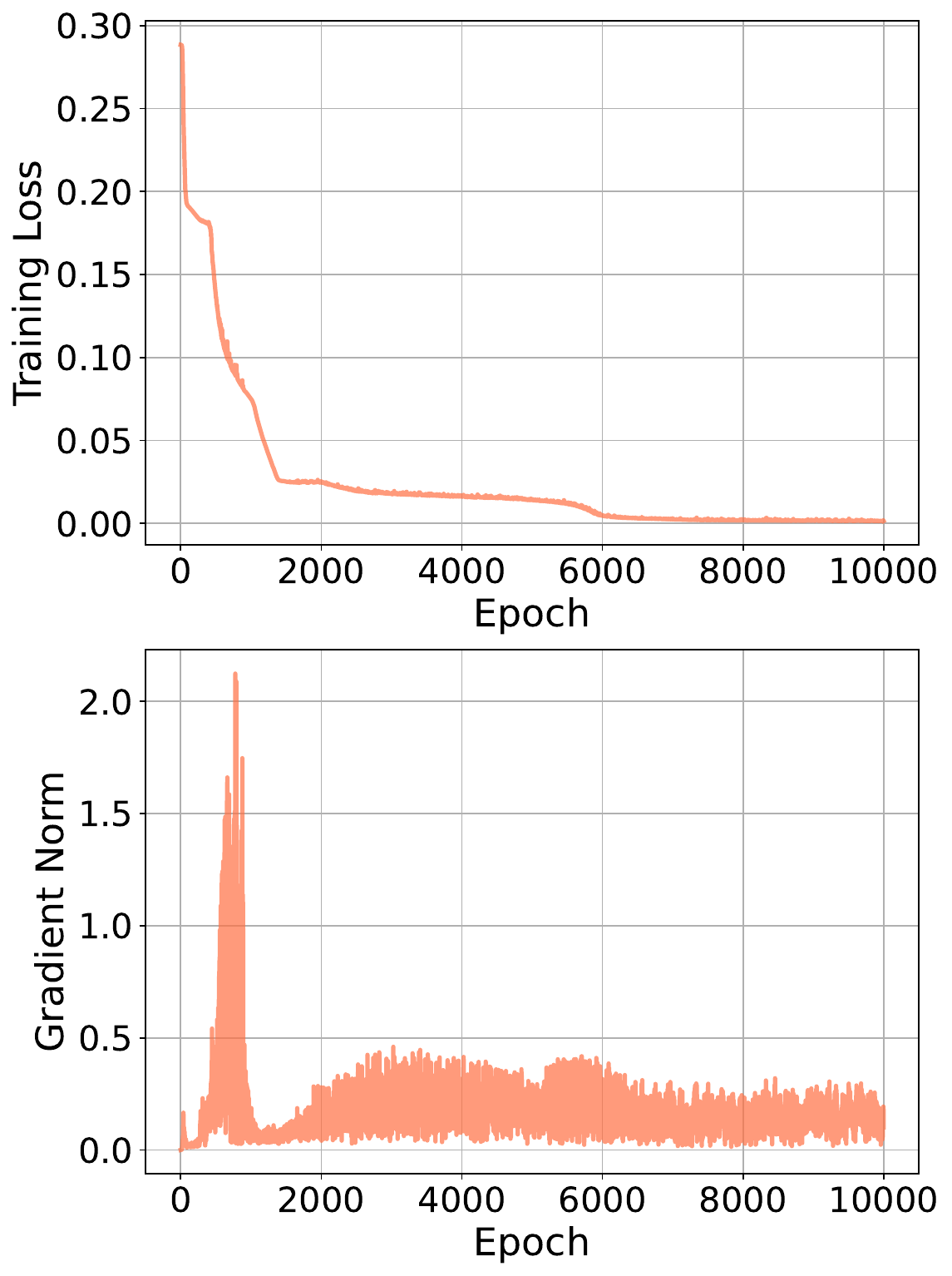}
    }\hspace{-3mm}
    \subfigure[Dopri5]{
    \includegraphics[trim={0 0 -23 0}, width=0.38\linewidth]{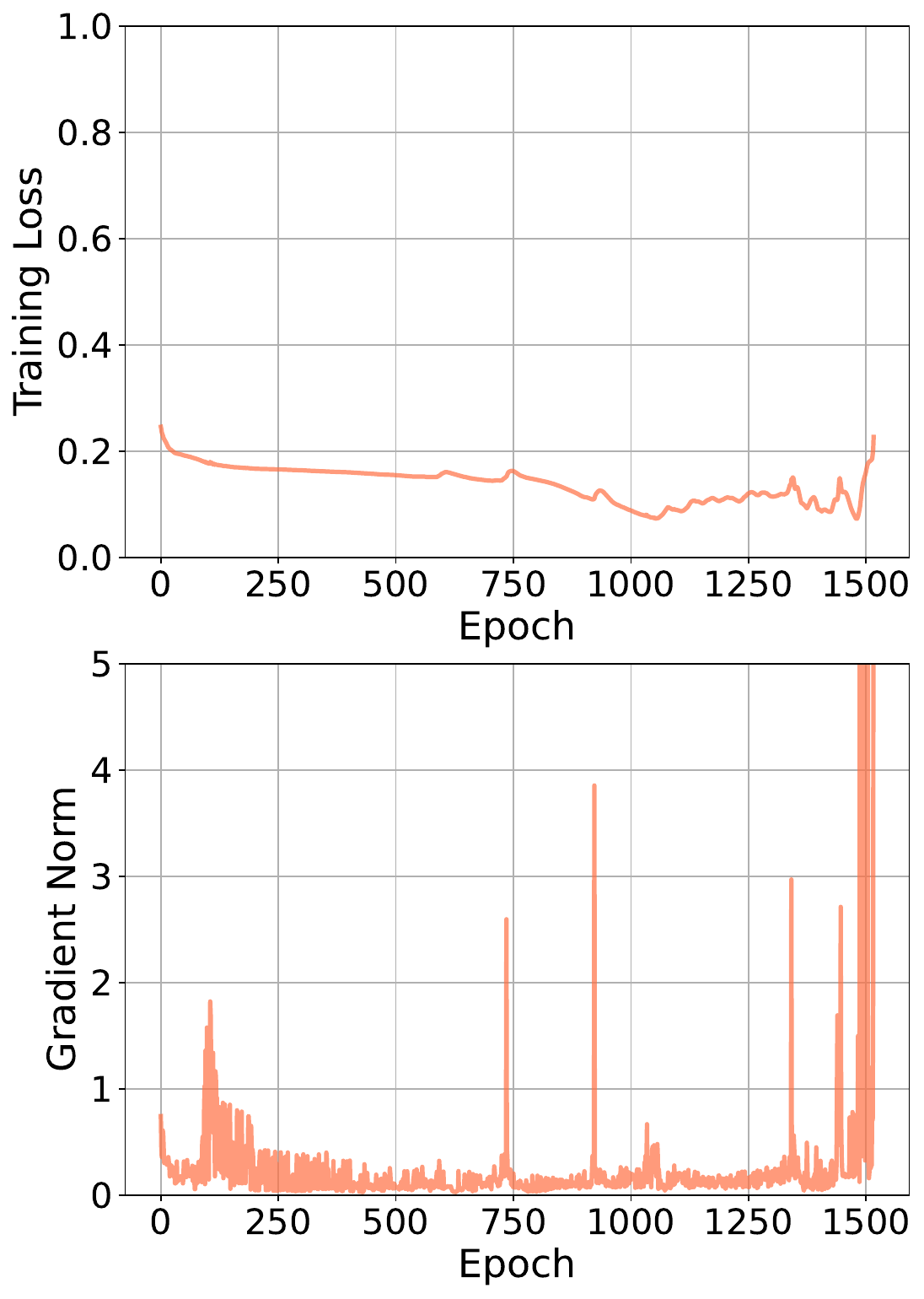}
    }
    \caption{Training results on Robertson's equations with the Crank--Nicolson method (left) and with the Dopri5 method (right). Training loss is plotted on the top, and gradient norm is plotted on the bottom. The gradients for Dopri5 explode after about 1,500 epochs.}
    \label{fig:rober-efficiency}
\end{figure}

PNODE enables us to use an implicit method to integrate the forward model and
use its discrete adjoint to compute the gradients for training. Here, we
demonstrate the advantage of implicit methods by comparing them with the
adaptive explicit methods that are widely used in existing neural ODE frameworks
because of their efficiency and ease of implementation. For implicit methods, we
use the Crank--Nicolson (CN) scheme. The nonlinear systems arising at each time
step are solved with a Newton method, and the linear systems are solved with a
matrix-free GMRES method \cite{Saad1986}. For explicit methods we use the
adaptive Dopri5 method with tolerances $abstol=reltol=10^{-6}$. As shown in
Figure \ref{fig:rober-efficiency}, PNODE with CN can learn the dynamics
perfectly, while Dopri5 fails to match the ground truth. This result is expected
because of the limitations of explicit methods in handling stiff systems. Figure
\ref{fig:rober-efficiency} shows that when using Dorpi5, the gradient explodes,
preventing the convergence of the training loss.

\subsubsection{Computation cost}

The training costs for Dopri5 and CN are given in Table \ref{tab:rober}. We can
see that training with CN is slightly faster than training with Dopri5 for the
Robertson's equations. Note that function evaluations are required not only for
the time integration but also for Autograd to generate the Jacobian-vector
product and the transposed Jacobian-vector product, which are needed in the
adjoint calculation. In principle, explicit methods have lower per-step cost
than implicit methods have because they do not require solving linear systems.
In our case, however, they require a large number of steps according to Table
\ref{tab:rober}. Initially, the training model is not stiff, requiring
relatively few time steps when integrating the neural ODE. As the training model
approaches the ground truth, however, the stiffness increases, causing the
number of time steps to increase as well.
\begin{table}
    \caption{Computation cost comparison between Dopri5 and CN.}
    \label{tab:rober}
    \centering
    \begin{tabular}{cccc}
      \toprule
      \multirow{2}{*}{\begin{minipage}{0.5in}\center Integration method\end{minipage}} &
      \multirow{2}{*}{\begin{minipage}{0.6in}\center Average NFE-F\end{minipage}} & \multirow{2}{*}{\begin{minipage}{0.6in}\center Average NFE-B\end{minipage}} & \multirow{2}{*}{\begin{minipage}{0.8in}\center Average time per iteration (s)\end{minipage}} \\ \\
       \midrule
       CN & 505 & 136 & 0.709 \\
       Dopri5 & 805 & 805 & 0.778 \\
      \bottomrule
    \end{tabular}
\end{table}

\section{Conclusion}\label{sec:conclusion}

In this work we propose PNODE, a framework for neural ODEs based on high-level
discrete adjoint methods with checkpointing. We show that the discrete adjoints
derived from a time integration scheme can be cast as a high-level abstraction
of automatic differentiation and produce gradients exact to machine precision.
Adopting this approach, one can avoid backpropagating through an ODE solver or a
time integration procedure by directly using ML platforms such as
\texttt{TensorFlow} and \texttt{PyTorch} and can minimize the depth of the
computational graph for NN backpropagation, while guaranteeing reverse accuracy.
With high-level adjoint differentiation and the checkpointing technique, we
successfully reduce the memory cost of neural ODEs to $\mathcal{O}(N_t N_s) +
\mathcal{O}(N_l)$ when checkpointing all intermediate states (including ODE
solutions and stage vectors) and $\mathcal{O}(N_t) +\mathcal{O}(N_l)$ when
checkpointing solutions only, whereas a naive automatic differentiation approach
requires a memory cost of $\mathcal{O}(N_t N_s N_l)$. Extensive numerical
experiments on image classification and continuous normalizing flow problems
show that PNODE achieves the best memory efficiency and training speed among the
existing neural ODEs that are reverse-accurate. Furthermore, our high-level
adjoint method not only allows a balance between memory and computational costs
but also offers more flexibility in the solver design than traditional neural
ODEs have. We demonstrate that PNODE enables the application of implicit
integration methods, which offers more possibilities for stabilizing the
training of stiff dynamical systems and building implicit deep learning models
\cite{Ghaoui2021}. We have made PNODE freely available and believe that
accelerated memory-efficient neural ODEs will benefit a broad range of
artificial intelligence applications, especially for scientific machine learning
tasks such as the discovery of unknown physics.



\bibliographystyle{IEEEtran}
\bibliography{references}



\end{document}